\documentclass[conference,9pt]{IEEEtran} 
\usepackage[dvipdfmx]{graphicx}
\usepackage{url}
\usepackage{cite}
\usepackage[cmex10]{amsmath}
\usepackage{amssymb}
\usepackage{bm}
\usepackage{color}
\usepackage{comment}
\usepackage{algorithm}
\usepackage{hyperref}

\newcommand{\argmin}{\mathop{\rm arg~min}\limits}
\newcommand{\mymin}{\mathop{\rm min}\limits}

\begin{document}
\title{
  \textcolor{black}{
    Automatic Generation of Typographic Font from a Small Font Subset
  }
}

\author{
  Tomo~Miyazaki,~\IEEEmembership{Member,~IEEE,}
  Tatsunori~Tsuchiya,
  Yoshihiro~Sugaya,~\IEEEmembership{Member,~IEEE,}
  Shinichiro~Omachi,~\IEEEmembership{Senior~Member,~IEEE,}
  Masakazu~Iwamura,~\IEEEmembership{Member,~IEEE,}
  Seiichi~Uchida,~\IEEEmembership{Member,~IEEE,}
  and~Koichi~Kise,~\IEEEmembership{Associate~Member,~IEEE}
  \thanks{Manuscript created September 25, 2015. This work was partially supported by JST, CREST, and JSPS KAKENHI Grant Numbers 16H02841 and 16K00259.}
  \thanks{T.~Miyazaki, Y.~Sugaya, and S.~Omachi are with Tohoku University~
    (e-mail: tomo@iic.ecei.tohoku.ac.jp).}
  \thanks{T.~Tsuchiya was with Tohoku University and is now with Canon, Inc.}
  \thanks{M.~Iwamura and K.~Kise are with Osaka Prefecture University, Japan.}
  \thanks{S.~Uchida is with Kyushu University, Japan.}
}

\maketitle

\begin{abstract}
  \textcolor{black}{
    This paper addresses the automatic generation of a typographic font from a subset of characters.
    Specifically, we use a subset of a typographic font to extrapolate additional characters.
    Consequently, we obtain a complete font containing a number of characters sufficient for daily use.
  }
  The automated generation of Japanese fonts is in high demand because a Japanese font requires over 1,000 characters.
  Unfortunately, professional typographers create most fonts,
  resulting in significant financial and time investments for font generation.
  The proposed method can be a great aid for font creation
  because designers do not need to create the majority of the characters for a new font.
  The proposed method uses strokes from given samples for font generation.
  The strokes, from which we construct characters, are extracted by exploiting a character skeleton dataset.
  This study makes three main contributions: a novel method of extracting strokes from characters,
  which is applicable to both standard fonts and their variations;
  a fully automated approach for constructing characters; and a selection method for sample characters.
  We demonstrate our proposed method by generating 2,965 characters in 47 fonts.
  Objective and subjective evaluations verify that the generated characters are similar to handmade characters.
\end{abstract}


\section{Introduction} \label{sec:intro}
\IEEEPARstart{A}{utomatic} generation of Japanese font is important for lowering the costs associated with creating kanji\footnote{Kanji are Chinese characters adapted for the Japanese language.} characters.
All of the characters in a font need to be designed in a particular style and size,
a process generally performed by humans.
Moreover, font creation is professional and time-consuming work.

The cost problem is especially crucial in languages that contain a very large number of characters,
such as Chinese, Korean, and Japanese.
Most commercial fonts contain at least 1,006 kanji characters,
which are defined as those in daily use by the Japanese Industrial Standards Committee\footnote{\url{http://www.jisc.go.jp}}.
Chinese and Korean each use more than 6,000 characters.
These numbers are far greater than those of other major languages,
e.g., 26 characters in English, 27 in Spanish, 49 in Hindi, 28 in Arabic, and 33 in Russian.
Thus, it takes a few years even for a professional designer to create a font.

\textcolor{black}{
Previous work on character generation is shown in Table \ref{tab:categorization}.
We categorized previous studies by their inputs and outputs into three problem settings.
The first problem setting is to generate a font using parameters
for handwritten \cite{id6,id7,id10,id15,id35,id23} or typographic fonts \cite{id33,id34,id37,id38,id20,id40}.
The second problem setting is to generate a font by blending several complete fonts\footnote{A complete font contains a sufficient number of characters for daily use, e.g., 26 for English and over a thousand for Japanese.}
for handwritten \cite{id17,id36,id45} or typographic fonts \cite{id60,id62,id53,id66}.
The third problem setting is to generate a font by extrapolating characters
given a subset of handwritten \cite{id46,id58,id47,id50,id18,id42,id51,id54} or typographic fonts \cite{id63}.
Each problem setting has been an important research topic.
}

\begin{table*}[t] \centering
  \textcolor{black}{
    \caption{Categories of related work}
    \label{tab:categorization}
  }
  \includegraphics{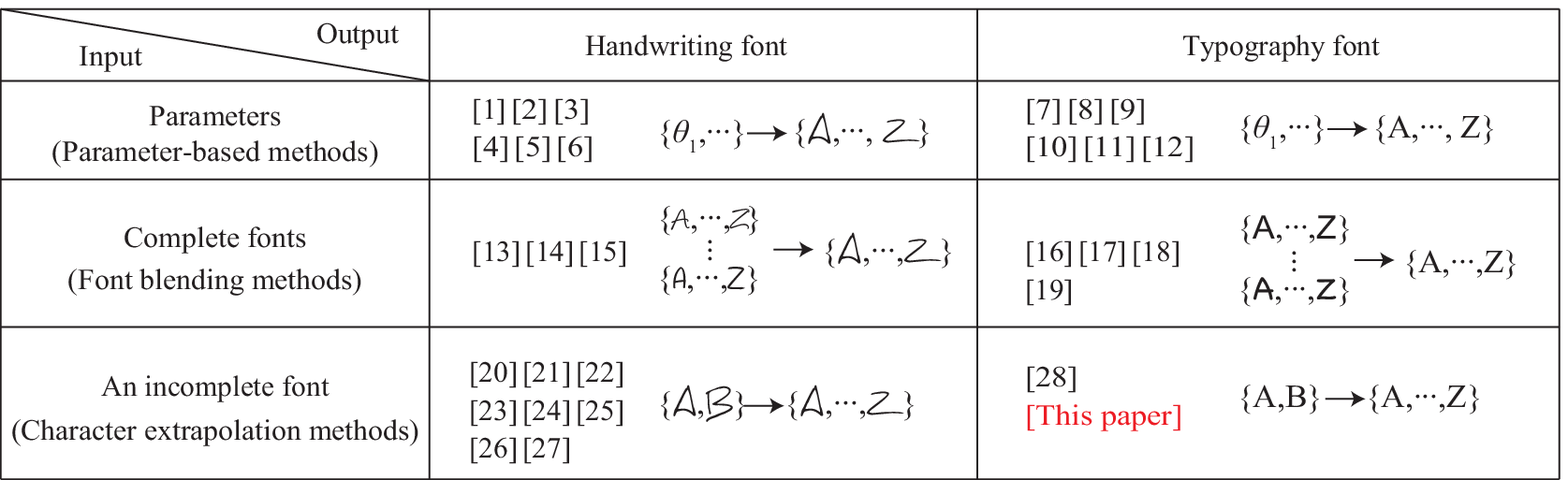}
\end{table*}

\begin{figure*}[t] \centering
  \includegraphics[width=7.16in]{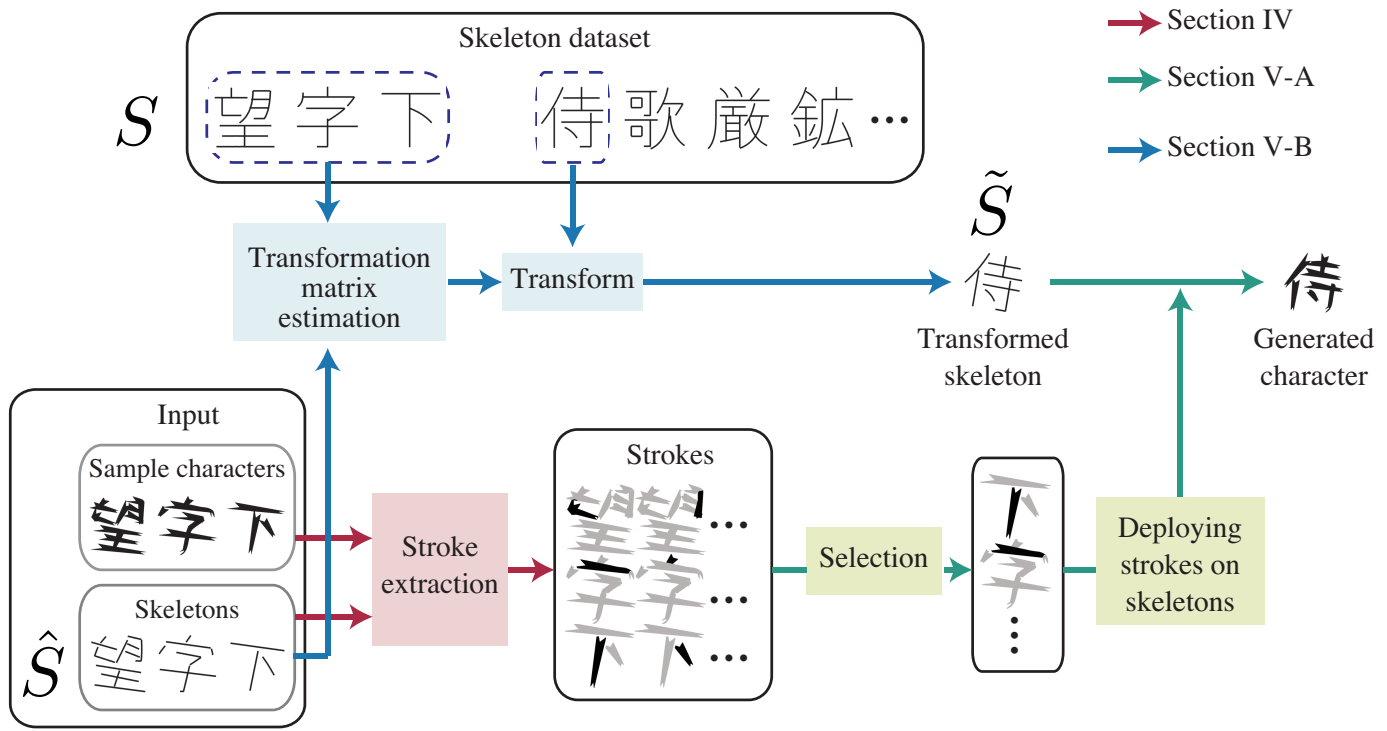}
  \caption{Overview of the proposed method.} \label{fig:overview}
\end{figure*}

\textcolor{black}{
  In this study, we address the problem of generating a typographic font from a given subset of characters.
  From a subset of a typographic font that includes only a small number of characters,
  we generate missing characters to obtain a complete typographic font.
  Although the proposed method requires a subset of a particular typographic font,
  the burden of creating a subset of a font is much less than creating an entire font.
  This benefit is highlighted in languages containing many characters, such as Chinese and Japanese.
}
The proposed method uses sample characters to extract strokes and constructs characters by deploying them.
Fig. \ref{fig:overview} illustrates an overview of the proposed method,
which begins with stroke extraction from sample characters.
Strokes are extracted using the skeletons of the samples.
We take the skeletons of a target character from the skeleton dataset
and transform it into the structure of the target font.
Finally, we select some strokes and deploy them to the target skeletons.

The proposed method accepts samples in an image format.
For practicality, the ideal font generation method must accept characters in an image format
and require only a small subset of the font as input.
First, we consider the image format. For the purpose of character recognition,
character generation from samples in an image format is far more useful than from those in a vector format.
Image formats are more convenient for sample collection.
For example, when users encounter text in an unknown font, it is difficult to find samples in a vector format,
whereas an image is available immediately upon taking a photo of the text.
Likewise, sample collection is time-consuming work
and could be an obstacle for the popularization of a font generation method if too many samples are needed.

Our automatic generation framework makes three main contributions,
each of which is briefly introduced below and described in detail in subsequent sections.

The first contribution of this study is a stroke extraction method. 
We propose a method that extracts character strokes
by applying a novel active contour model that determines the boundaries of strokes to extract.
Some strokes are damaged because of the complexity of characters;
therefore, we incorporate a restoration process for fixing these damaged strokes.

The second contribution of this study is a method
for the automatic construction of characters in two phases:
modifying skeletons in the target font and deploying strokes to them.
The proposed method extracts the transformation parameters.
With these parameters, we can unify the skeletons in size and geometry.
Because our method transforms skeletons from a standard font (such as Mincho) to fit the target font,
it is unnecessary to build skeleton datasets for a large number of fonts. 
This approach helps us with variety and scalability.
We can generate as many characters as the skeletons in the dataset will allow
(210,000 characters are available at this time).
For automation, we define an energy function to deploy a stroke
so that an appropriate one can be added to the skeleton.

The third contribution of this study is a sample selection method.
We address which characters are suitable for the proposed method
and how we can select them based on the energy function used in character generation.
Finding samples is combinatorial problem, and it is difficult to obtain a solution analytically.
Therefore, we apply a genetic algorithm to find an approximate solution.

\textcolor{black}{
  \section{Related work} \label{sec:related}
}
In this section we review the relevant work illustrated in Table \ref{tab:categorization}. 

\subsubsection*{Parameter-based methods}
The methods in this category use parameters acquired by analyzing fonts.
One of the most famous methods is Metafont\cite{id33},
which use the shapes of pen tips and pen-movement paths as parameters for character generation.
\cite{id34} was inspired by Metafont.
Further, \cite{id37, id38} analyzed characters in the Times font family to extract parameters.
Character properties, such as thickness and aspect ratio, are changed using these parameters.
A method for generating Japanese characters was addressed by Tanaka {\it et al.} \cite{id20},
who created a database of skeletons of Japanese characters, including kanji.
The characters were generated by placing thick lines in the skeleton data.
A famous noncommercial font\footnote{\url{https://www.tanaka.ecc.u-tokyo.ac.jp/ktanaka/Font/}}
was created by applying this method.
Kamichi extended the skeleton database in \cite{id40}.

There have been attempts to analyzing handwritten characters.
Bayoudh {\it et al.} used Freeman chain-code to analyze alphabetical characters \cite{id6}.
Djioua and Plamondon used a Sigma lognormal model supported by the kinematic theory \cite{id7}.
An interactive system was developed so that the user can easily fit a Sigma-lognormal model to alphabetical characters.
Wada {\it et al.} extracted the trajectories of alphabetical characters
and replaced them using a genetic algorithm \cite{id10}.
Zheng and Doermann adopted a thin plate spline to model an alphabetical character
and generated a new alphabetical character by calculating the intermediate of the two \cite{id15}.
Handwriting models for robot arms were developed in \cite{id35, id23}. 

Parameter-based methods generate clean characters.
However, they are only applicable to particular fonts that are analyzed by humans.
For instance, the method in \cite{id20} accepts only two fonts: Mincho and Gothic.
This drawback comes from the method's requirement for precise analysis.
In addition, the methods cannot be applied to Japanese fonts,
since over a thousand of characters need to be parameterized.
Moreover, special equipment is required to accurately parameterize alphabetical characters,
such as an interactive system or device for acquiring trajectories.
Unlike these methods, the method proposed in this study generates characters without requiring precise analysis.
Thus, only a few character sample images are required.
As we discussed in the previous section, preparing sample images is a simple task.

\subsubsection*{Font blending methods}
Methods in this category receive several complete fonts and generate a font by blending them.
Xu {\it et al.} generated Chinese calligraphy characters
using a weighted blend of strokes in different styles \cite{id17}.
They decomposed samples into radicals and single strokes based on rules defined by expert knowledge.
An improved method \cite{id36} considers the spatial relationship of strokes.
Choi {\it et al.} generated handwriting Hangul characters using a Bayesian network trained with a given font\cite{id45}.

Suveeranont and Igarashi addressed a generation of alphabetical characters for typographic fonts \cite{id60}.
They generated characters by blending predefined characters from miscellaneous complete fonts.
This method is based on a vector format, in contrast with the proposed method, which accepts an image format.
Campbell and Kautz learned a manifold of standard fonts of alphabetical characters \cite{id62}.
Locations on the manifold represent a new font.
Feng {\it et al.} used a wavelet transform to blend two fonts \cite{id53}.
Tenenbaum and Freeman used complete fonts as references to form characters from the generation results \cite{id66}. 

The methods in this approach require human supervision to generate a desired font;
since they automatically blend fonts, the result is not always the desired font.
In order to avoid this problem, blending weights \cite{id17} or an interactive system \cite{id60} are implemented.
However, sweeping the weights and using the system require human supervision.

In contrast, the proposed method generates a typographic font with little human supervision.
Instead of blending fonts,
we construct characters with strokes extracted from sample characters in the desired font.
Therefore, the proposed method can directly provide the user with the desired font.

\subsubsection*{Character extrapolation methods}
The aim of methods in this category is to extrapolate characters not included in a given subset.
Attempts to complete this process on handwritten fonts can be found in \cite{id46,id58,id47,id50,id18,id42,id51,id54}.
Lin {\it et al.} generated characters with components extracted from given characters \cite{id46}
using an annotated font in which the positions and sizes of components were labeled.
The extraction of the components was performed on electronic devices
so that characters can be easily decomposed.
Zong and Zhu developed a character generation method using machine learning \cite{id58}.
They decomposed the given characters into components by analyzing the orientation of strokes.
Components were assigned to a reference font with a similarity function trained by a semi-supervised algorithm.
Wang {\it et al.} focused on the spatial relationships of the character components
for decomposition and generation \cite{id47}.
An active contour model was used for decomposition \cite{id50}.

Character component decomposition is a crucial technique for the methods in this category.
However, most use naive decomposition that rely on
spatial relationships \cite{id50,id47,id42,id51,id18} or special devices \cite{id46,id54}.
It is difficult to extract components when they are connected or decorated.
To the best of our knowledge, the method in \cite{id63} is the only method
that is applicable to characters in fonts with decorations.
Saito {\it et al.} applied a patch transform \cite{cho2008patch} to samples
and generated alphabetical characters in wide range of fonts \cite{id63}.
However, the generated results did not meet the criteria for practical use.
In this paper, we propose an adaptive active contour model for component extraction.
With the proposed method, we can obtain natural character strokes even in decorated characters.

\section{Skeleton dataset} \label{sec:dataset}
\begin{figure}[t] \begin{center}
    \begin{tabular}{cc}
      \includegraphics[width=1in]{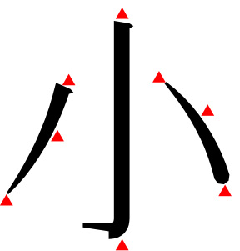}
      &\includegraphics[width=1in]{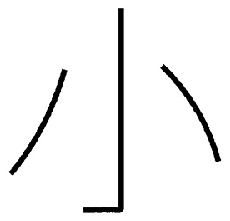} \\ (a) &(b)
    \end{tabular}
    \caption{
      Visualization of the GlyphWiki data and our dataset.
      (a) Data from GlyphWiki drawn in Mincho.
      The red triangles represent the control points.
      The line types of the left and right strokes are ``curve'' and ``straight'' in the middle.
      All start and end shapes are different.
      (b) The skeletons in our dataset.
      The control points, line type, start shape, and end shape are attributes.
    } \label{fig:data}
\end{center} \end{figure}

First, we address the character skeleton dataset used in this study.
We created the dataset based on GlyphWiki \cite{glyphwiki},
which contains data from more than 210,000 of characters\footnote{It includes characters created by users that are not used publicly.}.
GlyphWiki utilizes a wiki format, allowing anybody to contribute to and maintain the database.
Each stroke of a character is stored in KAGE format \cite{id40} which uses four attributes: control points, line type, start shape, and end shape.
There are six line types, seven start shapes, and 15 end shapes.
The number of control points depends on line type and is at most four.
The control points and line types lead to a rough stroke,
then the start and end shapes provide details of a stroke.
Fig. \ref{fig:data} (a) illustrates a character in KAGE format in Mincho.

We created skeletons of the strokes in a character using GlyphWiki data.
Fig. \ref{fig:data}(b) illustrates the skeletons of a character in our dataset.
We draw a line of a skeleton from the first control point to the last control point.
An additional line of a skeleton is drawn according to start shape and end shape.
For example, the end shape is drawn in the middle stroke in Fig. \ref{fig:data}(b).
We draw a line according to the line type and the number of control points.
A straight line is drawn when the line type is straight and two control points are given.
A B-spline curve is drawn when the line type is curved and three control points are given.
A straight line and B-spline curve are drawn
when the line type is vertical and four control points are given.
We include the attributes with each skeleton.
We extract points on a skeleton by regular sampling from the start point to the end point.
Therefore, a skeleton is composed of a set of sampling points. 
By using sampling points, we define a skeleton $S$ as
\begin{equation}
  S = \left[ \begin{array}{ccc}
      \cdots & x_i & \cdots \\
      \cdots & y_i & \cdots \\
      \cdots & 1 & \cdots
    \end{array} \right],
\end{equation}
where $x_i$ and $y_i$ represent the $x$- and $y$-coordinates of $i$th sampling point, respectively.
We adopted homogeneous coordinates.
$P = \left[ x, y, 1 \right]^\mathrm{T}$ denotes a vector that represents a sampling point.
Then, $S = \left[ \cdots, P_i, \cdots \right]$.
We denote a skeleton in our dataset by $S$,
a skeleton in the samples by $\hat{S}$,
and a transformed skeleton\footnote{We described the details of the transformed skeletion in Section \ref{sec:skeleton_modification}.} by $\tilde{S}$ (See Fig. \ref{fig:overview}).

\section{Stroke extraction} \label{sec:extraction}
\begin{figure}[t] \begin{center}
    \includegraphics[width=2.5in]{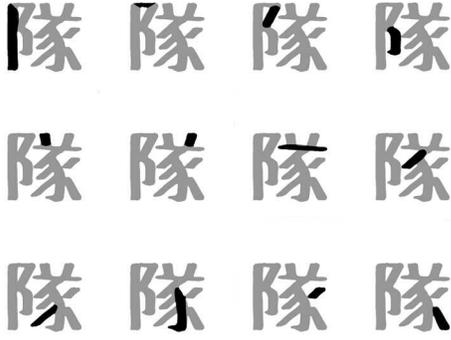}
    \caption{Stroke extraction.
      The black pixels represent the extracted strokes.} \label{fig:parts}
\end{center} \end{figure}

When the proposed method receives samples as an input, we extract the strokes of the samples.
Fig. \ref{fig:parts} shows examples of extracted strokes.

\subsection{Adjusting skeletons to samples}
The proposed method maximally exploits the skeletons of sample characters,
however, the skeletons of samples are not given.
The skeletons can be easily extracted if strokes are separated; however, strokes often overlap.
Quality skeleton extraction is essential for the proposed method.
Therefore, inspired by accurate segmentation applications, such as Grabcut \cite{Rother:2004},
we adopt user interaction to adjust skeletons in our dataset to the samples
so that accurate skeletons of the samples can be obtained.

\begin{figure}[t]
  \begin{center} \begin{tabular}{cc}
      \includegraphics[width=1in]{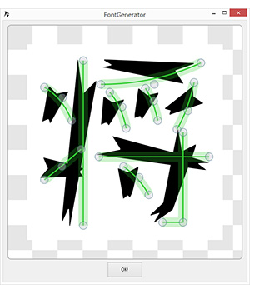}
      &\includegraphics[width=1in]{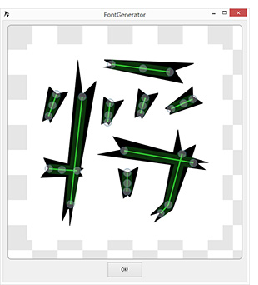}\\
      (a) &(b)
    \end{tabular}
    \caption{
      Skeleton adjustment GUI.
      (a) Screenshot of the GUI initiating an adjustment.
      Note that the skeleton is not matched to the sample at this point
      because the skeletons are in Mincho.
      (b) Adjustment result.
    } \label{fig:modification}
  \end{center}
\end{figure}

We developed a graphical user interface (GUI) for adjusting skeletons,
as illustrated in Fig. \ref{fig:modification}.
Skeletons are displayed with control points on the sample.
The operator adjusts a skeleton by dragging the control points
and the GUI shows these changes in real time.

In order to assist the operator, we implemented three automation techniques:
scale adjustment, rotation adjustment, and cooperative move.
Firstly, a scale of skeletons is adjusted to fit the sample.
We calculate a scaling factor from rectangles around the skeleton and sample.
The second technique is rotation adjustment.
We extract points from the skeleton by sampling.
Likewise, we extract points from the medial axis of the sample,
which is obtained by a morphological operation.
With these points, we use iterative closest point matching \cite{Geiger:2012}
to calculate the rotation matrix.
Thirdly, control points move cooperatively when the start point is moved by hand,
e.g., if the start point moves up, the other control points also move up.
The scale and rotation adjustments are performed only once, before manual adjustment.

We emphasize that skeleton adjustment is not difficult and can be completed in minutes.
It is unnecessary to adjust all of the skeletons in the dataset.
According to our experimental results,
approximately 15 samples are sufficient for the proposed method.
Therefore, skeleton adjustment is significantly less work
than creating thousands of characters by hand.

\subsection{Skeleton relation assignment} \label{sec:relation}
\begin{figure}[t] \begin{center}
    \includegraphics[width=3in]{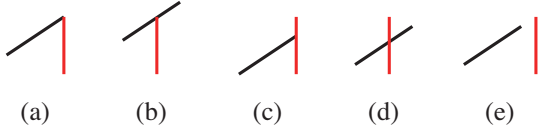}
    \caption{
      Skeleton relations of the red skeletons to the black skeletons:
      (a) continuous, (b) connecting, (c) connected, (d) crossing, and (e) isolated.
    } \label{fig:states}
\end{center} \end{figure}

We determine the connectivity of the skeletons in the samples.
We define relations between two skeletons as follows:
\begin{itemize}
\item continuous: the start or end of a skeleton is connected to the start or end of another.
\item connecting: the start or end of a skeleton is connected to the body of another.
\item connected: the body of a skeleton is connected to the start or end of another.
\item crossing: the skeleton crosses over another.
\item isolated: the skeleton is isolated from others.
\end{itemize}
The body is the part of a skeleton without start or end parts.
Examples of each relation are illustrated in Fig. \ref{fig:states}.

\begin{figure}[t] \begin{center}
    \includegraphics[width=1.4in]{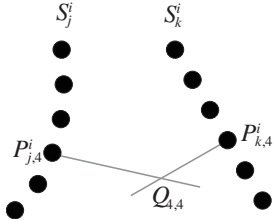}
    \caption{
      Measurement of the distance between two skeletons.
      The sampling points are drawn with black circles.
    } \label{fig:Pjk}
\end{center} \end{figure}

\begin{figure}[t] \begin{center}
    \begin{tabular}{cc}
      \includegraphics[width=1in]{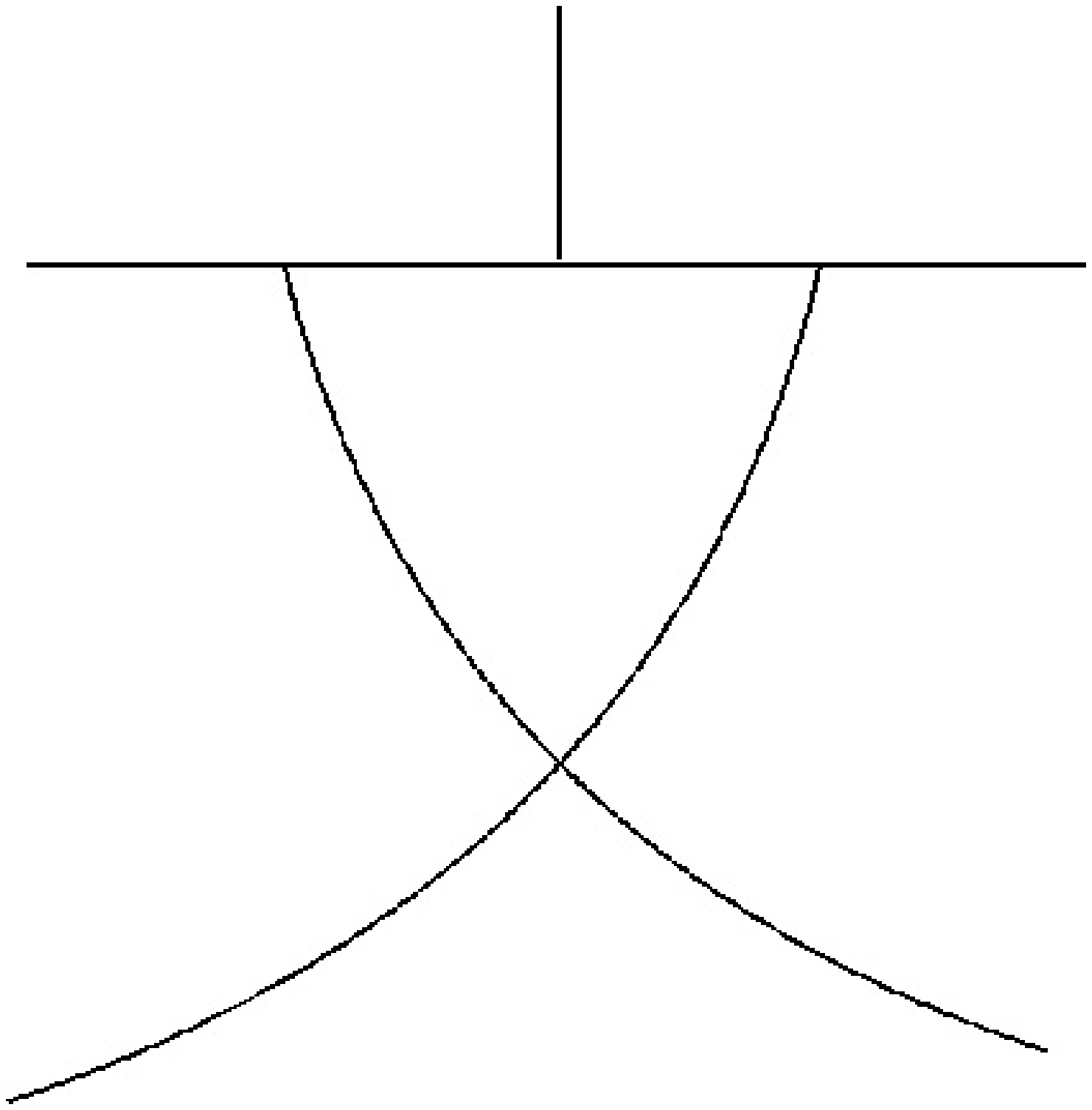}
      &\includegraphics[width=1in]{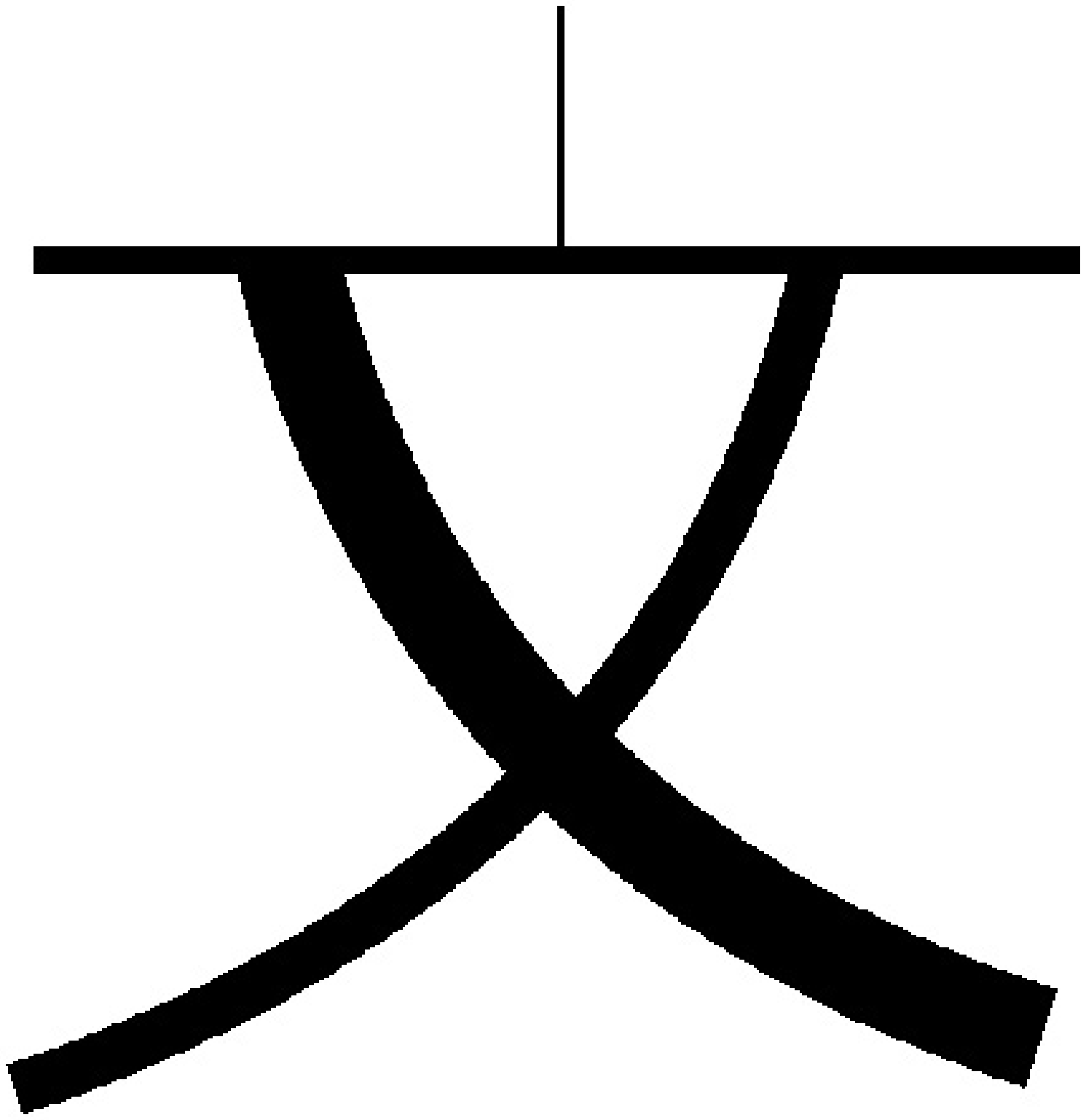}\\
      (a) &(b)
    \end{tabular}
    \caption{(a) Skeletons and (b) created characters with $\tau = 1, 10, 20, 40$.} \label{fig:tau} 
\end{center} \end{figure}

Two skeletons are in contact when the distance between them is small.
Therefore, we measure the distance $d$ between $S$ and $S'$ as
\begin{equation}
  d = \left \{ \begin{array}{ll}
    \mymin_{i,j} \| Q_{i,j} - P_i \|_2 + \| Q_{i,j} - P'_j \|_2 &\mbox{if $Q$ exists},\\
    \infty &\mbox{otherwise}. \end{array} \right . \label{eq:distance}
\end{equation}
$Q_{i,j}$ is the point at which two lines cross,
which are vertical to $S$ and $S'$ and through $P_i$ and $P'_j$, respectively.
Fig. \ref{fig:Pjk} illustrates an example of $Q$.
Using the $\bar{i}$ and $\bar{j}$ that minimize $d$,
we can determine which part of $S$ is in contact with $S'$,
e.g., the start of $S$ is in contact with the part of $S'$
when $\bar{i}$ is small and $\bar{j}$ is large.
We use three functions to indicate contact: $\mathbb{I}_s$ for the start,
$\mathbb{I}_e$ for the end, and $\mathbb{I}_b$ for the body.
For $n$ points of a skeleton, the functions calculate true or false, as follows.
$\mathbb{I}_s(\bar{i})=\text{True}$ if $\frac{\bar{i}}{n}<.05$, otherwise False.
$\mathbb{I}_e(\bar{i})=\text{True}$ if $\frac{\bar{i}}{n}>.95$, otherwise False.
$\mathbb{I}_b(\bar{i})=\text{True}$ if $.05 \le \frac{\bar{i}}{n} \le .95$, otherwise False.
The parameters $.05$ and $.95$ were determined experimentally.
We assign a relation $R$ to $S$ against $S'$ as follows:
\begin{equation}
  R = \left \{ \begin{array}{ll}
    \mbox{continuous} &\mbox{if } d < 2(\tau+\tau') \\
    & \quad \wedge \left( \mathbb{I}_s(\bar{i}) \vee \mathbb{I}_e(\bar{i}) \right) \\
    & \quad \wedge \left( \mathbb{I}_s(\bar{j}) \vee \mathbb{I}_e(\bar{j}) \right), \\
    \mbox{connecting} & \mbox{if } d < 2(\tau+\tau') \\
    & \quad \wedge \left( \mathbb{I}_s(\bar{i})
    \vee \mathbb{I}_e(\bar{i}) \right) \wedge \mathbb{I}_b(\bar{j}), \\
    \mbox{connected} & \mbox{if } d < 2(\tau+\tau') \\
    & \quad \wedge \mathbb{I}_b(\bar{i}) \wedge \left( \mathbb{I}_s(\bar{j})
    \vee \mathbb{I}_e(\bar{j}) \right), \\
    \mbox{crossing} & \mbox{if } d < 2(\tau+\tau') \\
    & \quad \wedge \mathbb{I}_b(\bar{i}) \wedge \mathbb{I}_b(\bar{j}), \\
    \mbox{isolated} &\mbox{otherwise},
  \end{array} \right . \label{eq:strokeStates}
\end{equation}
where $\tau$ represents a thickness of a stroke in a sample.
We can draw a character image with $\tau$.
Fig. \ref{fig:tau} illustrates the skeletons and created character with various values of $\tau$.
We empirically determined $\tau$ by creating character images
with various $\tau$ and choosing the $\tau$ that minimizes the hamming distance
between the created and sample character images.

We reconsider the relations obtained above.
Since the relations are based on the skeletons,
it is necessary to verify that they are consistent with the sample images.
The relations may occasionally be different from the samples
because of human error during adjustment and the parameters.
The parameter $2(\tau+\tau')$ in Eq. (\ref{eq:strokeStates}) is set to a severe value
in order to detect all stroke contacts without omission.
Consequently, an incorrect relation may be assigned.

\begin{figure}[t] \begin{center}
    \begin{tabular}{cc}
      \includegraphics[width=1in]{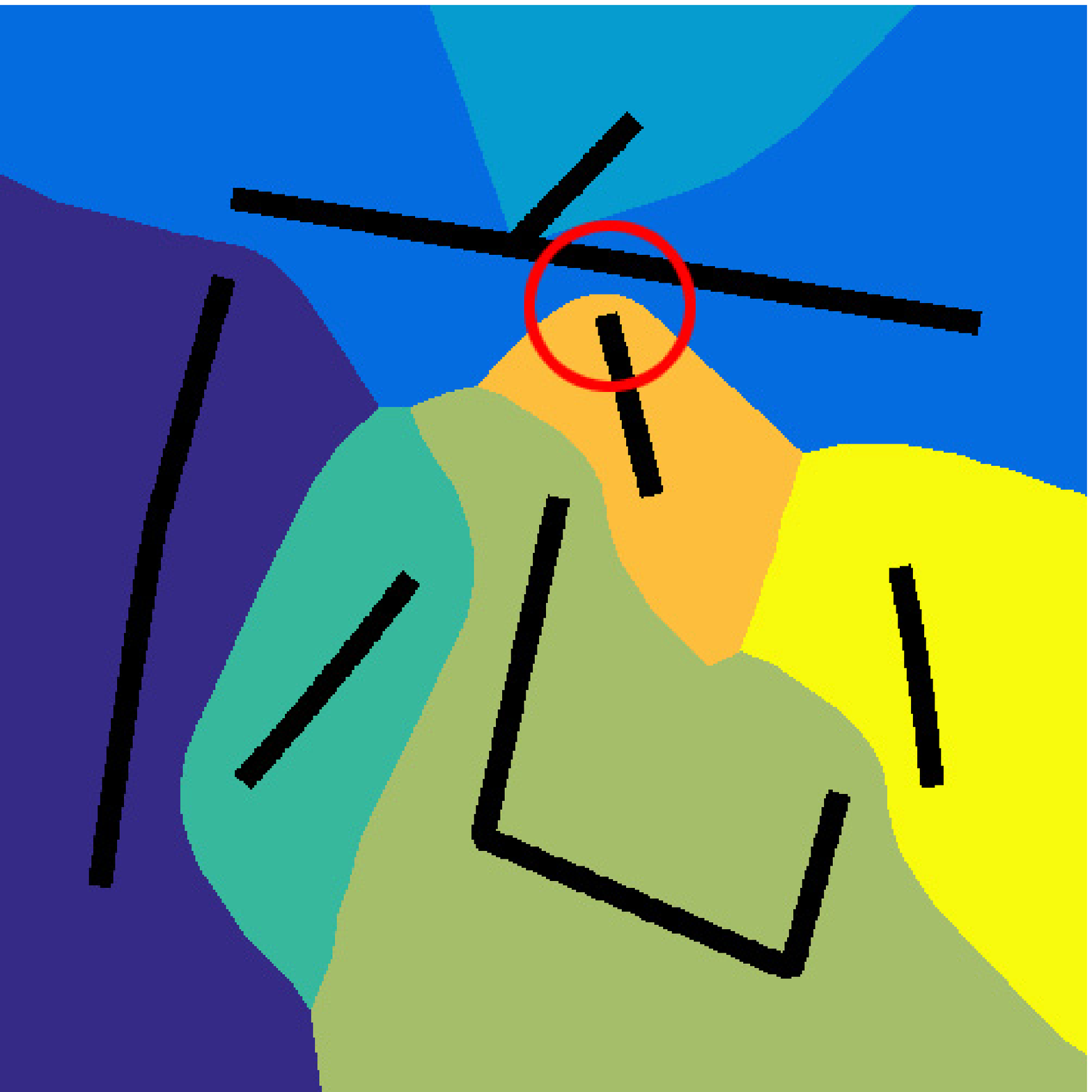}
      &\includegraphics[width=1in]{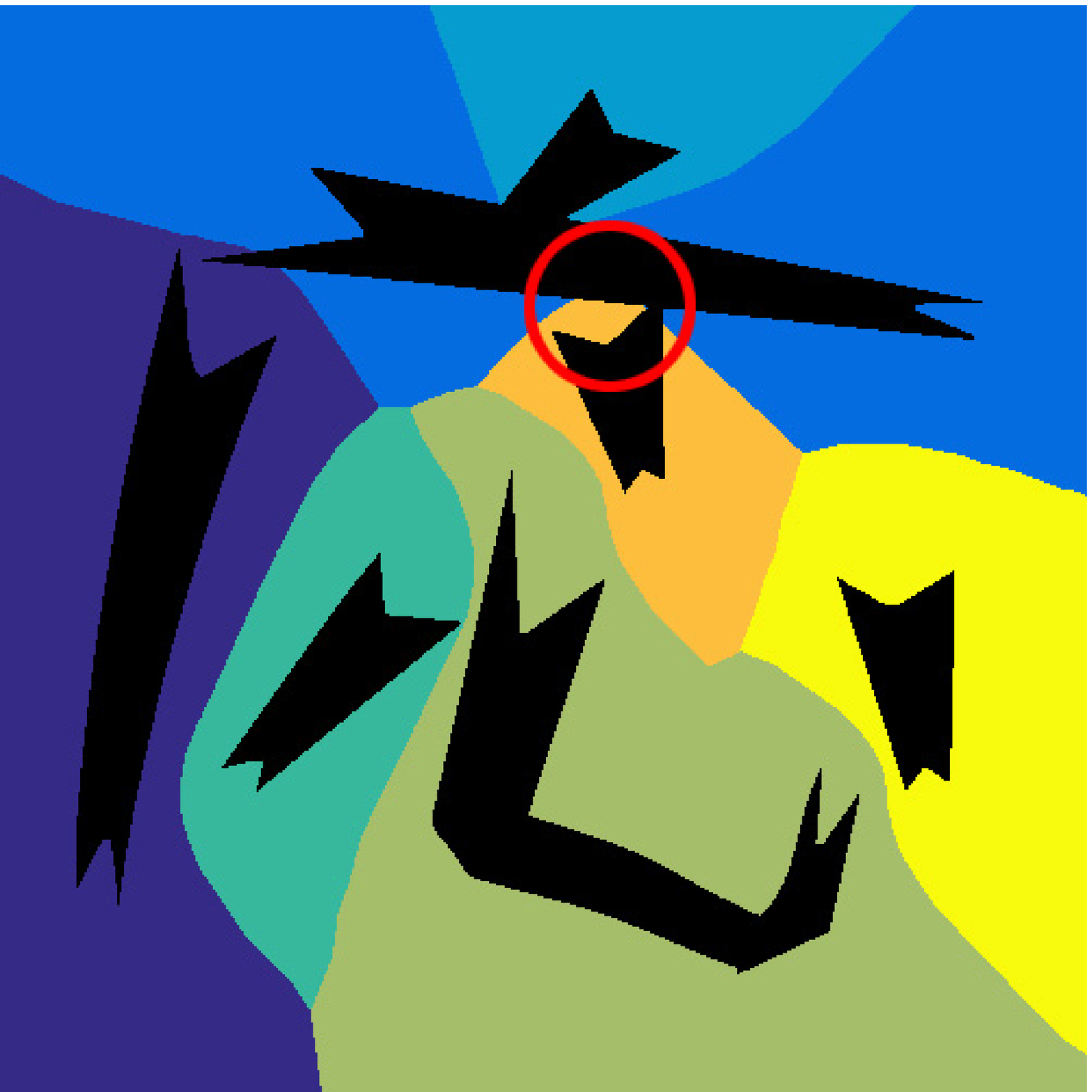}\\ (a) &(b)
    \end{tabular}
    \caption{
      Relation check.
      (a) Skeleton of segmentation results.
      (b) Sample of the segmentation results.
      Focusing on the two segments indicated by the red circle,
      the relation is isolated in (a), but there is only one connected component in (b).
    } \label{fig:segmentation}
\end{center} \end{figure}

The procedures for reconsideration are as follows.
We classify all pixels of the sample to the nearest skeleton within the Euclidean distance.
Fig. \ref{fig:segmentation} (a) illustrates the segmentation results.
Subsequently, we choose a connected component from the sample using these results.
Then, we determine the number of connected components obtained.
If the relation is isolated, there must be two connected components.
Likewise, if the relation is connecting, connected, or crossing,
there must be only one connected component.
This check complements Eq. (\ref{eq:strokeStates}).

\subsection{Adaptive active contour model}
We propose an adaptive active contour model (AACM) for character stroke extraction.
The AACM is able to distinguish even strokes that have contact with others,
determining the boundary of each stroke.
Inspired by Snake \cite{kass1988snakes}, we seek the boundary by optimizing a spline curve.
In addition, we incorporate constraints and adaptive energy modification into the AACM
so that strokes can be extracted from complex characters.

The AACM is defined as a spline curve that minimizes the energy function as
\begin{equation}
  E_{\text{\it AACM}} = E_{\text{\it int}} + E_{\text{\it img}},
\end{equation}
where $E_{\text{\it int}}$ and $E_{\text{\it img}}$ take into account the smoothness of the curves
and the fit to objects, respectively.
$E_{\text{\it int}}$ follows \cite{kass1988snakes}.
$E_{\text{\it img}}$ must be determined for each task.
If we set $E_{\text{\it img}}$ to $-|\triangle I|$,
where $\triangle I$ represents the gradient image of $I$, $E_{\text{\it img}}$ pulls the AACMs to the edges.
For our purposes, it is unnecessary for a boundary to be very close to its stroke,
and instead, may be rather relaxed.
If the boundary is too close, it may cross the stroke. 
Consequently, the extracted strokes are damaged.
Therefore, in this paper, we define $E_{\text{\it img}}$ as
\begin{equation}
  E_{\text{\it img}} = G_v \ast I_{\text{\it SMP}} + I_{\text{\it SK}}, \label{eq:Eimg}
\end{equation}
where $G_v$ is a Gaussian kernel with variance $v$, and $I_{\text{\it SMP}}$ and $I_{\text{\it SK}}$ are
the grayscale images\footnote{Pixel values are 0 if they are background, and 255 otherwise.} of
the sample and its skeletons, respectively.
$E_{\text{\it img}}$ has a high
energy\footnote{The pixel value represents the energy: 0 is the lowest and 255 is the highest.}
around the strokes, thus, the boundary remains separated from them.
We use the boundaries of the segmentation results as the initial AACMs.
We adaptively modify $E_{\text{\it img}}$ and set some points that the AACMs must pass through.

\begin{figure}[t] \begin{center}
    \begin{tabular}{cc}
      \includegraphics[width=1in]{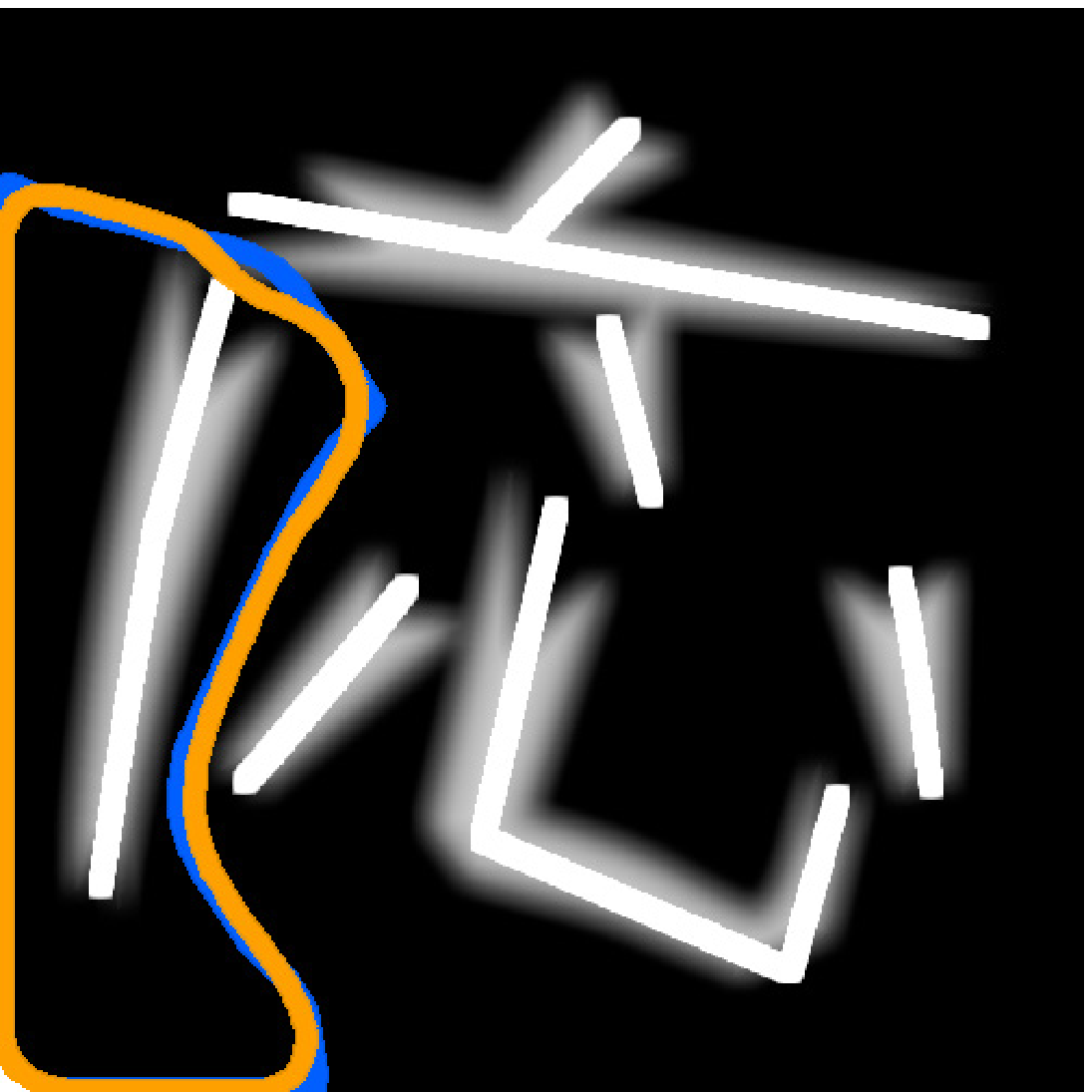}
      &\includegraphics[width=1in]{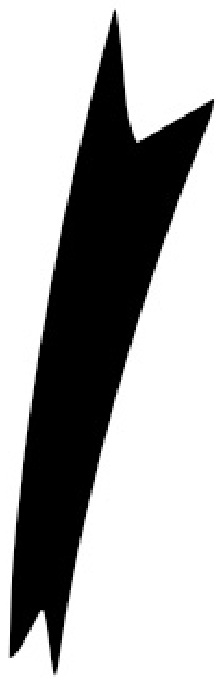}\\
      (a) &(b)
    \end{tabular}
    \caption{
      Extraction of an isolated stroke.
      (a) Initial AACM (blue) and minimized AACM (orange) on $E_{\text{\it img}}$.
      (b) Extracted stroke.
    } \label{fig:snake_isolate}
\end{center} \end{figure}

In the case where the relation of the target stroke is isolated,
which is the simplest case, we directly apply an AACM.
Fig. \ref{fig:snake_isolate} illustrates the results of the extraction of an isolated stroke.

\begin{figure}[t] \begin{center}
    \begin{tabular}{cc}
      \includegraphics[width=1in]{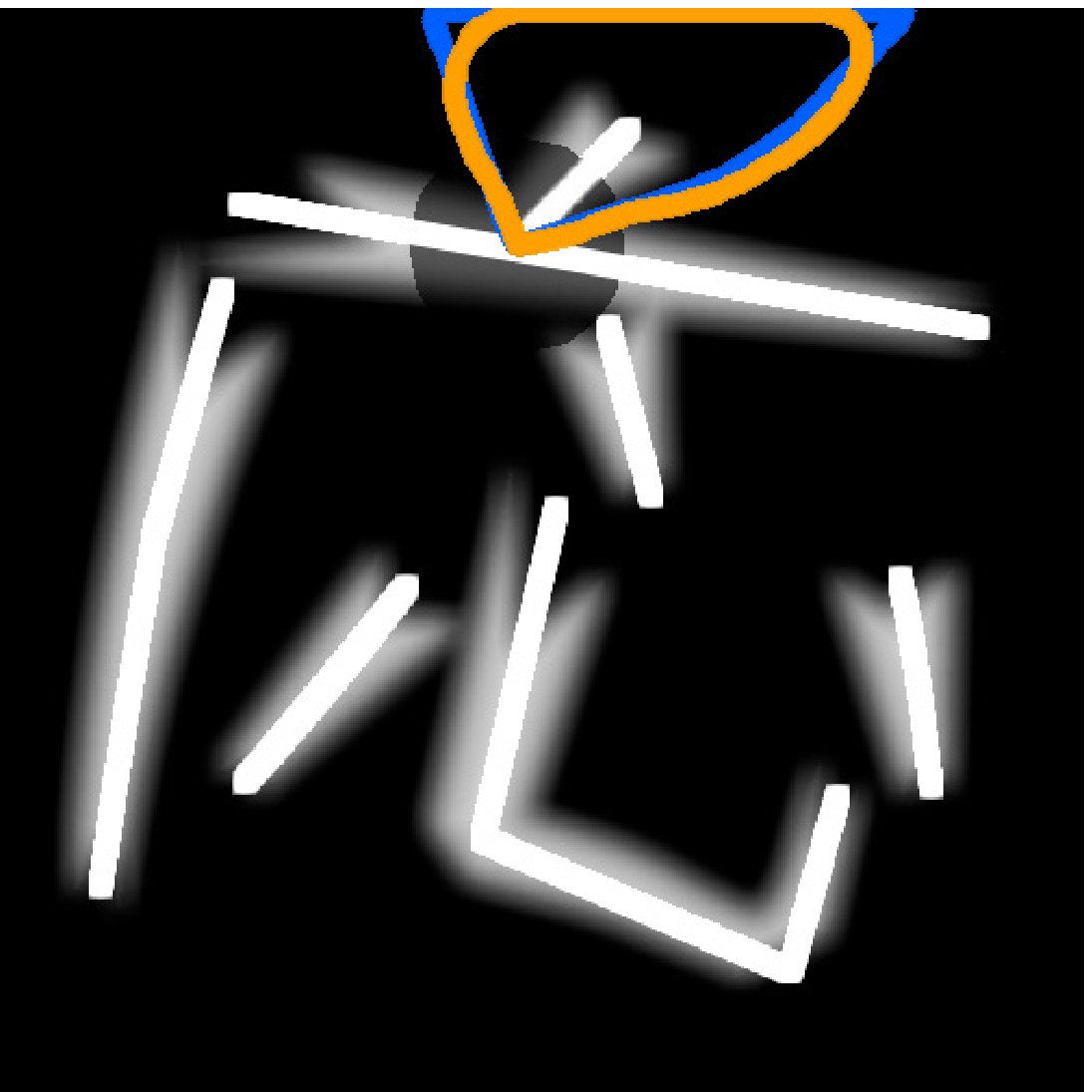}
      &\includegraphics[width=1in]{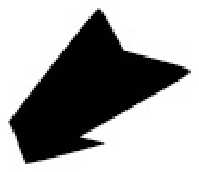}\\
      (a) &(b)
    \end{tabular}
    \caption{
      Stroke extraction for connecting or continuous stroke relations.
      (a) Initial AACM (blue) and minimized AACM (orange) on $E_{\text{\it img}}$.
      (b) Extracted stroke.
    } \label{fig:snake_connecting}
\end{center} \end{figure}

In the case where the relation of the target stroke is connecting or continuous,
we force an AACM to go through the point at which the strokes are in contact.
Then, we decrease $E_{\text{\it img}}$ by $1/\beta_1$
within the range of the length $\beta_2 \tau$ from the point,
except for $I_{\text{\it SK}}$.
This decrease operation encourages the AACM to pass through the stroke intersection.
$\beta_1$ and $\beta_2$ are constant values that were determined empirically.
With the modified $E_{\text{\it img}}$ and the constraints,
we minimize the AACM and extract the target.
Fig. \ref{fig:snake_connecting} demonstrates the extraction for this case.
The AACM naturally passes through the intersection due to the decreased energy.

Where the relation of the target stroke is crossing or connected,
we decrease $E_{\text{\it img}}$ by $1/\beta_1$
within the range of $\beta_2 \tau_{\text{\it others}}$ from the other strokes.
Then, we minimize the AACM with the modified $E_{\text{\it img}}$
and extract the target along the minimized AACM.

\subsection{Stroke restoration}
\begin{figure}[t] \begin{center}
    \begin{tabular}{cc}
      \includegraphics[width=1in]{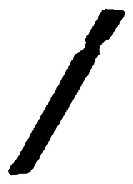}
      &\includegraphics[width=1in]{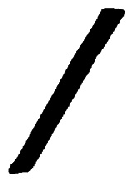} \\ (a) &(b)\\
      \includegraphics[width=1in]{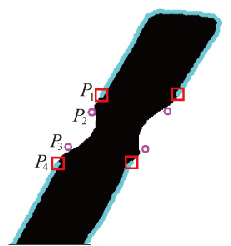}
      &\includegraphics[width=1in]{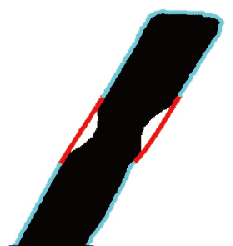} \\ (c) &(d)\\
    \end{tabular}
    \caption{
      Restoration of an extracted stroke.
      (a) Damaged stroke. (b) Restored stroke.
      (c) Contours removed with cubic B\'ezier curves;
      red rectangles represent $P_1$ and $P_4$,
      and pink circles represent $P_2$ and $P_3$.
      Cyan lines represent contours.
      (d) The created curves, indicated by red lines.
    } \label{fig:restoration}
\end{center} \end{figure}

We reshape extracted strokes,
which often have defects at points where contact occurs; see Fig. \ref{fig:restoration}(a).
An intuitive method for restoration is the removal of contours around defects
and the connection with cubic B\'ezier curves.
A cubic B\'ezier curve is drawn with four points $(P_1,P_2,P_3,P_4)$.
The curve starts from $P_1$ and ends at $P_4$. $P_2$ and $P_3$ serve as control points
that provide a direction to the curve.

\begin{figure}[t] \begin{center}
    \includegraphics[width=3.2in]{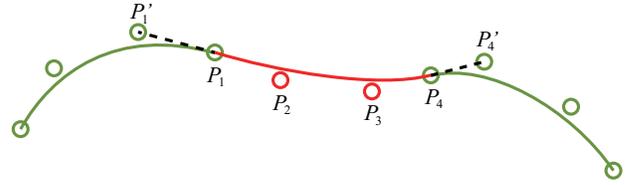}
    \caption{
      Points of a new B\'ezier curve.
      The red curve represents a new curve and the green curves represent those remaining.
    } \label{fig:points}
\end{center} \end{figure}

We apply cubic B\'ezier curves to a stroke using POTRACE \cite{selinger2003potrace},
generating a number of curves that represent the short parts of the contour.
Then, we remove the curves within a distance of $\tau$ from the extracted stroke;
see Fig. \ref{fig:restoration}(c).
We create a cubic B\'ezier curve between two curves so that the contour can be continuous;
see Fig. \ref{fig:restoration}(d).
We set $P_1$ and $P_4$ of the new curve to $P_4$ and $P_1$, respectively, of the remaining curves.
Fig. \ref{fig:points} illustrates the new curve.
We calculate $P_2$ and $P_3$ by
\begin{eqnarray}
  P_2 &=& P_1 + \gamma (P_1 - P'_1),\\
  P_3 &=& P_4 + \gamma (P_4 - P'_4),
\end{eqnarray}
where $\gamma$ is a constant value.
$P'_1$ and $P'_4$ represent the nearest control points from $P_1$ and $P_4$ in the remaining curves, respectively.
$P_1 - P'_1$ and $P_4 - P'_4$ represent local gradients.
Hence, $P_2$ and $P_3$ are the points moved along the local gradients from $P_1$ and $P_4$, resulting in a smooth curve.

\section{Stroke deployment} \label{sec:generation}
We generate characters by deploying strokes on skeletons.
Our character generation method consists of two phases:
skeleton modification and stroke deployment.

\subsection{Skeleton modification} \label{sec:skeleton_modification}
The results of character generation would be strange---even with perfect strokes---
if the style of the skeleton dataset were different from the target font.
Thus, we modify the skeletons to be similar to the target font.
A feasible method for modification is the use of a transformation matrix.
We estimate the transformation matrix from the skeletons of the samples.
Specifically, we seek two transformation matrices: $T_{\text{\it sz}}$ and $T_{\text{\it aff}}$.
$T_{\text{\it sz}}$ adjusts the size of the skeletons and centroid translation,
and $T_{\text{\it aff}}$ adjusts the affine transformation of the skeletons.
Modification is carried out by applying $T_{\text{\it sz}}$ and $T_{\text{\it aff}}$ to the dataset.

We estimate $T_{\text{\it sz}}$ from the rectangles of the skeletons.
Let $H$ and $W$ denote the height and width of a character in the dataset, respectively.
Likewise, let $\hat{H}$ and $\hat{W}$ be the height and width of the skeleton of a sample character, respectively.
With output image size $I_w$ and $I_h$,
we calculate $T_{\text{\it sz}}$ by averaging size transformation
over the sample characters $\bm{C}=\{ \cdots, c_i, \cdots \}$ as: 
\begin{equation}
  T_{\text{\it sz}} = \frac{1}{\left \vert{\bm{C}} \right \vert} \sum_{c_i \in \bm{C}} \left( \begin{array}{ccc}
    \frac { \hat{W}_{c_i} } { W_{c_i} } &0 & I_w\frac{\hat{W}_{c_i}}{2W_{c_i}} \\
    0 & \frac { \hat{H}_{c_i} } { H_{c_i} } &I_h\frac{\hat{H_{c_i}}}{2H_{c_i}} \\
    0 &0 &1
  \end{array} \right).
\end{equation}

\begin{figure}[t] \begin{center}
    \begin{tabular}{cccc}
      \includegraphics[width=0.7in]{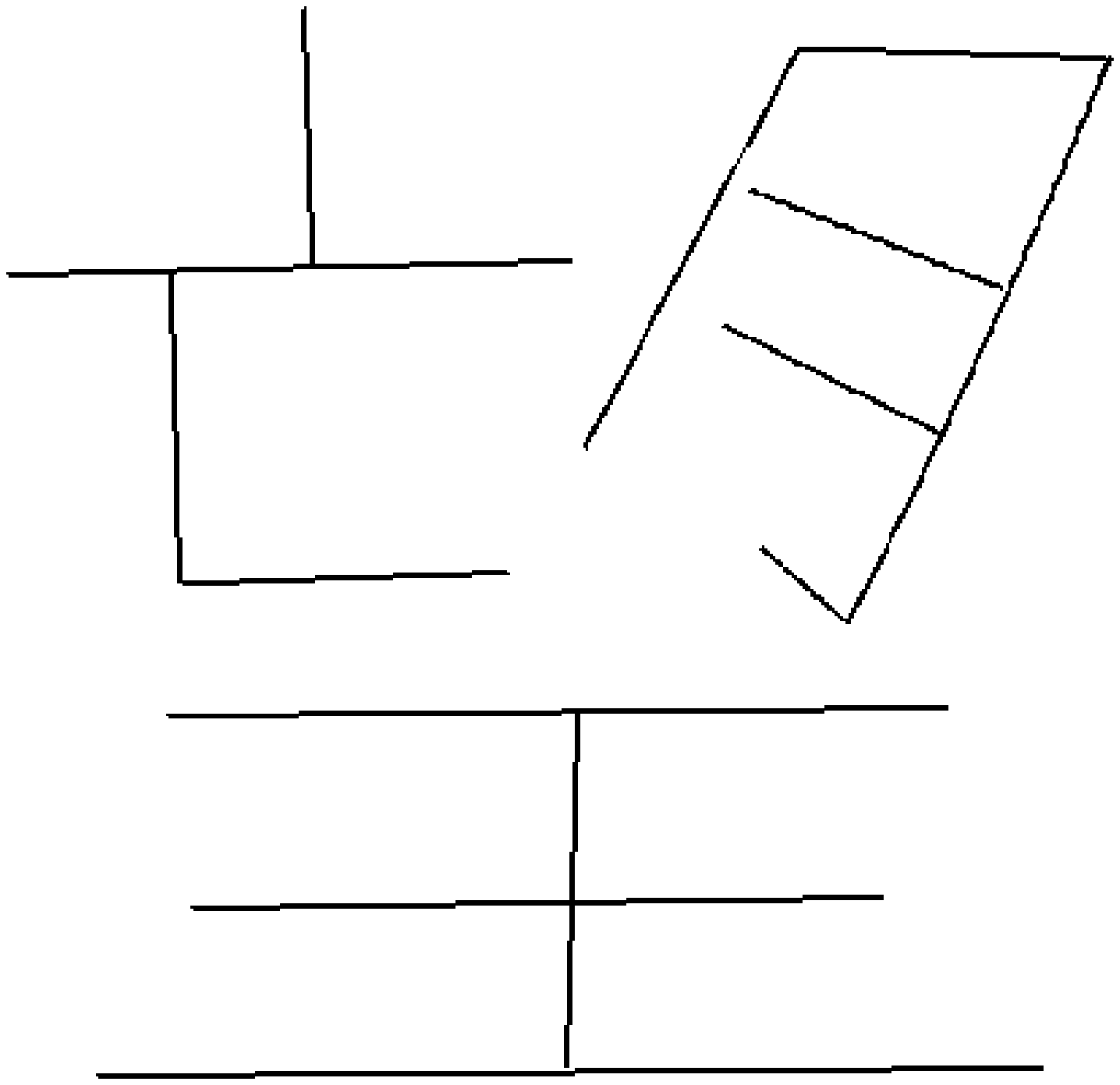}
      &\includegraphics[width=0.7in]{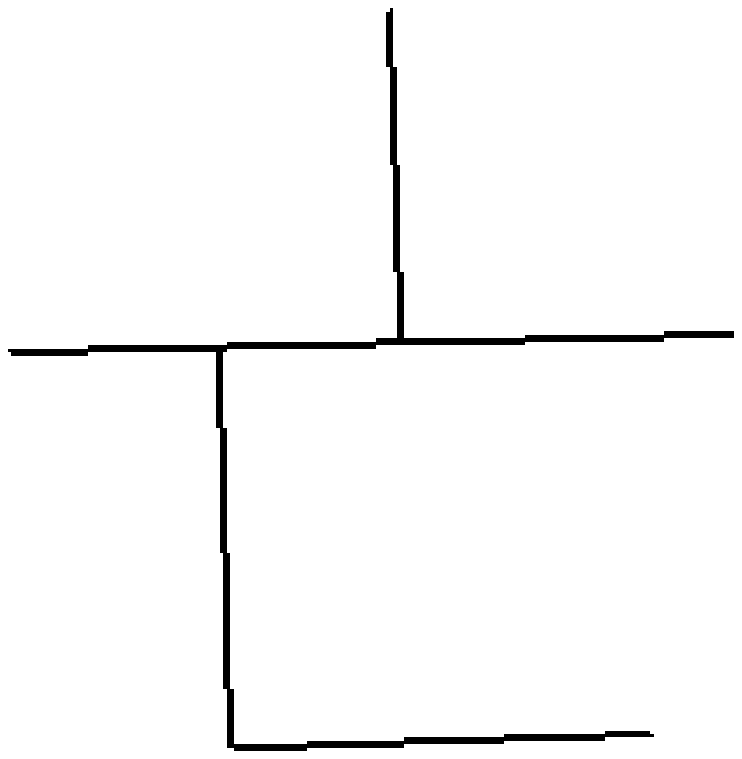}
      &\includegraphics[width=0.7in]{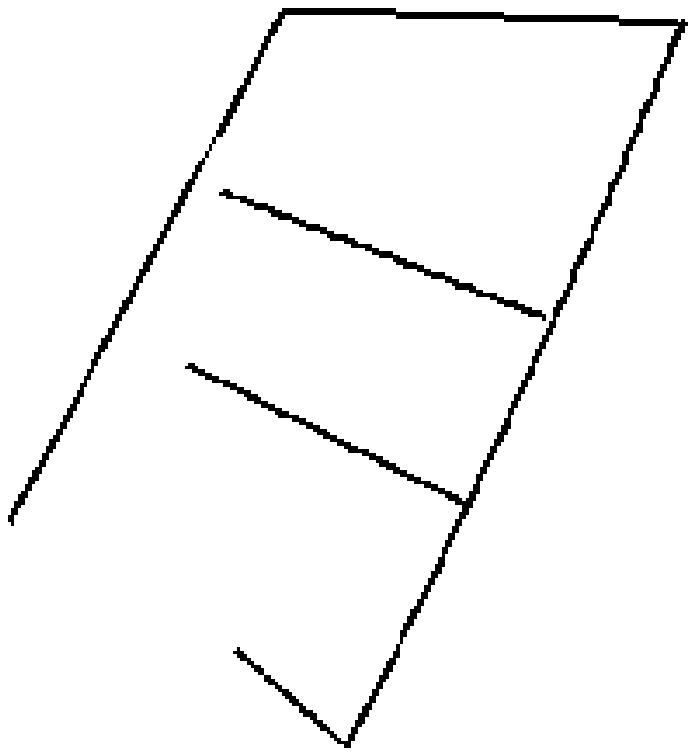}
      &\includegraphics[width=0.7in]{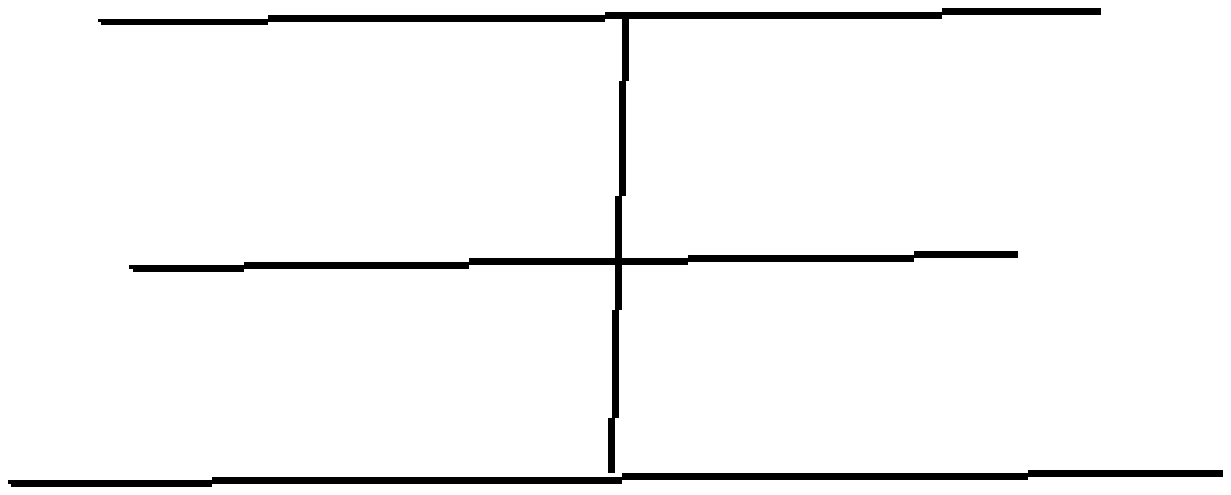}\\
      (a) &(b) &(c) &(d)
    \end{tabular}
    \caption{
      Groups of skeletons: (a) skeleton and (b), (c), and (d) groups.
    } \label{fig:group}
\end{center} \end{figure}

We estimate $T_{\text{\it \it aff}}$ using the skeletons of the sample and dataset.
However, it is difficult to estimate $T_{\text{\it aff}}$ directly from the skeletons
because of the complex structure of the characters.
If the characters are complex, they have many skeletons.
The affine transformations of skeletons cancel each other.
Eventually, $T_{\text{\it aff}}$ becomes a trivial matrix, such as an identity matrix.
To avoid these problems,
we divide the characters into groups and calculate $T_{\text{\it aff}}$
by averaging the affine transformation matrices obtained from each group.
We divide a character using the relations of skeletons.
Groups consist of skeletons whose relations are continuous, connecting, connected, and crossing.
Fig. \ref{fig:group} illustrates the grouping results.
We use function $f_T$ to calculate the affine transformation matrix,
which fits stroke $S$ to stroke $S'$.
We formulate $f_T$ based on the least-squares method, as Eq. (\ref{eq:fT_1}),
which can be solved analytically as Eq. (\ref{eq:fT_2}). 
\begin{eqnarray}
  f_T(S,S') &=& \argmin_{T \in \bm{T}} \| S - T S' \|_2, \label{eq:fT_1} \\
  &=& ( S^\top S ) ^{-1} S^\top S', \label{eq:fT_2}
\end{eqnarray}
where $\bm{T}$ represents a set of all possible affine transformation matrices.
We calculate $T_{\text{\it aff}}$ by averaging affine transformation matrices from $S$ to $\hat{S}$ as
\begin{equation}
  T_{\text{\it aff}} = \frac{1}{n_{\text{strokes}}} \sum_{i \in \bm{C}}
  \sum_{k=1}^{n^i_\text{groups}} \sum_{j \in G^i_k} f_T(S^i_j,\hat{S}^i_j), \label{eq:Ta}
\end{equation}
where $n_{\text{strokes}}$ is the number of total strokes in $\bm{C}$.
The number of groups in character $i$ is $n^i_\text{groups}$.
$G^i_k$ is a set of stroke indices in group $k$ of character $i$,
such as: $G^i_k = \{ j \mid S^i_j \in \text{group $k$} \}$.

Finally, we modify the skeletons by applying $T_{\text{\it sz}}$ and $T_{\text{\it aff}}$.
A transformed stroke is obtained using the equation
\begin{equation}
  \tilde{S} = T_{\text{\it aff}} T_{\text{\it sz}} S.
\end{equation}

\subsection{Stroke deployment}
Here, we describe the framework for selecting a stroke
that is the most suitable for a target skeleton $\tilde{S}$.
Stroke deployment has an impact on the appearance of the generated characters;
therefore, the appropriate stroke must be chosen for each target skeleton.

First, we select $\hat{S}$, which fits to $\tilde{S}$, from a set of the extracted strokes $\hat{\bm{S}}$.
Then, we determine the stroke $\hat{\mathcal{I}}$ corresponding to the selected $\hat{S}$.
\begin{eqnarray}
  \hat{\mathcal{I}} &=& f_{\text{\it img}} ( \hat{S} ),\\
  \hat{S} &=& \argmin_{ \hat{S}_i \in \hat{\bm{S}} } E(\hat{S}_i,\tilde{S}), 
\end{eqnarray}
where $f_{\text{\it img}}$ is a function that gives a stroke corresponding to $\hat{S}$
and $E$ is an energy function associated with skeletons $\hat{S}$ and $\tilde{S}$.
Once $\hat{\mathcal{I}}$ is determined,
we apply $f_T(\hat{S},\tilde{S})$ to $\hat{\mathcal{I}}$
and deploy the transformed $\hat{\mathcal{I}}$ on $\tilde{S}$.
We repeat the selection until strokes for all target skeletons are determined.
Finally, a character is generated by integrating the selected strokes.

We define $E$ as
\begin{eqnarray}
  E( \hat{S}, \tilde{S} ) = E_s( \hat{S}, \tilde{S} ) +
  E_s( f_{\text{\it data}}(\hat{S}), f_{\text{\it data}}(\tilde{S}) ) + \nonumber \\
  E_a( f_{\text{\it data}}(\hat{S}), f_{\text{\it data}}(\tilde{S}) ), \label{eq:energy}
\end{eqnarray}
where $f_{\text{\it data}}(S)$ gives the skeleton corresponding to $S$ in the skeleton dataset.
The energy $E$ utilizes three terms in inspecting $\hat{S}$.
The first term measures the distance between skeletons in a sample character and a target character.
When this term is small, two skeletons are similar.
Therefore, $\hat{\mathcal{I}}$ will naturally fit to the target skeleton.
The second term also measures the distance between two strokes, but only using the dataset.
If the skeletons in the dataset are similar, the stroke is favorable for the target skeleton.
This term improves the accuracy of the energy function.
The third term measures distance using adjacent skeletons,
making the distance more global than when focusing on only two skeletons.

The energy $E_s$ measures the distance between two skeletons.
We define $E_s(S,S')$ as
\begin{eqnarray}
  E_s( S, S' ) &=& 
  \| S' - f_T(S,S') S \|_1 \nonumber \\
  && + \| S - f_T(S',S) S' \|_1 \nonumber \\
  && + \sum_{k=1}^3 \mathbb{I}_k(S,S'),
\end{eqnarray}
where $\mathbb{I}_k$ is $1$ if the three attributes of two skeletons---line type,
start shape, and end shape---are different; otherwise, it is $0$.
$\mathbb{I}_k$ inspects the line type, start shape,
and end shape of the stroke when $k = 1$, $2$, and $3$, respectively.
The first term of $E_s$ measures the distance between two skeletons.
The second term of $E_s$ takes into account the distortion from the transformation $f_T(S,S')$;
a less distorted skeleton is favored.
The third term of $E_s$ incorporates attribute differences.
We define $E_a$ as
\begin{equation}
  E_a(S,S') = \begin{cases}
    \| S_\text{\it st} - S'_\text{\it st} \|_1 +
    \| S_\text{\it ed} - S'_\text{\it ed} \|_1
    & \text{if } \bm{S}_a \text{exists}\\
    50 & \text{otherwise},
  \end{cases}
\end{equation}
where $\bm{S}_a = \{ S_\text{\it st}, S_\text{\it ed}, S'_\text{\it st}, S'_\text{\it ed} \}$,
and $S_{\text{\it st}}$ represent a skeleton connected to the start point of $S$.
In the case where there are several skeletons connected to $S$ at the start point,
we choose one skeleton whose center is closest to $S$.
Likewise, $S_{\text{\it ed}}$ represents a skeleton connected to the end point of $S$.
If $S_{\text{\it st}}$, $S_{\text{\it ed}}$, $S'_{\text{\it st}}$,
or $S'_{\text{\it ed}}$ is unfavorable, $E_a$ has a large value.

\section{Sample selection} \label{sec:sample}
The proposed method uses $\bm{C}$, a font subset, to generate a large number of characters.
The generation results are deeply influenced by $\bm{C}$. 
Therefore, it is important to analyze which characters are suitable elements for $\bm{C}$.
With an optimal $\bm{C}$, we are able to maximize the capability of the proposed method.
We define a process of seeking the optimal $\bm{C}$ as {\it sample selection}.

In sample selection, we use the skeleton dataset to seek $\bm{C}$.
Note that sample images are not used.
Since the skeletons in the dataset are fundamental data,
if the skeletons in $\bm{C}$ in the dataset are suitable,
$\bm{C}$ in a target font can be expected to work well.

In order to ensure sample selection feasibility,
we seek $\bm{C}$ in a subset of the dataset that we define as {\it validation characters}.
In this study, we adopt the 1,006 characters of kyoiku-kanji containing the elemental characters of Japanese.
Since the number of characters in the dataset exceeds 210,000, it is time-consuming to use the entire dataset.
As we show in experimental results in Section \ref{sec:experiments},
the proposed method is able to generate acceptable results with a selected $\bm{C}$,
verifying the effectiveness of our approach.

Sample selection is based on a genetic algorithm.
There are a large number of candidates of optimal $\bm{C}$ even with the subset.
Candidates undergo crossover, evaluation and selection processes and
an optimal candidate can be obtained over many iterations of these processes.

We define function $f_{\text{\it selection}}$
that calculates the energy of a candidate $\bm{C}_i$
using a set of skeletons $\bm{S}_i$ of $\bm{C}_i$.
Let $\bm{S}_v$ represent a set of skeletons of validation characters. 
\begin{eqnarray}
  f_{\text{\it selection}}(\bm{S}_i) &=& \alpha f_e(\bm{S}_i) + (1-\alpha) f_r(\bm{S}_i), \\
  f_e(\bm{S}_i) &=& \sum_{s_v \in \bm{S}_v} \left ( \min_{s \in \bm{S}_i} E_s(s,s_v) + E_a(s,s_v) \right),  \\
  f_r(\bm{S}_i) &=& |\bm{S}_i| + N_{\text{\it cross}}(\bm{S}_i) + N_{\text{\it con}}(\bm{S}_i),|N|
\end{eqnarray}
where $|\bm{S}|$ represents the cardinarity.
$N_{\text{\it cross}}$ gives the number of skeletons whose relations are crossing.
$N_{\text{\it con}}$ gives the number of skeletons whose relations are continuous, connecting, and connected.
$f_r$ measures the complexity of $\bm{C}$ and serves as a regularization term.
We control the complexity of the samples with a constant value $\alpha$.

\textcolor{black}{
  \section{Experimental results} \label{sec:experiments}
} 
We demonstrate character generation with the proposed method using kanji.
We used five fonts as the target fonts: Ibara \ref{font:ibara}), Gcomic \ref{font:gcomic}),
Onryo \ref{font:onryo}), Tsunoda \ref{font:srtsunoda}), and Zinpen \ref{font:zinpen}), where the indices represent the numbers in Appendix.
Fig. \ref{fig:examples} shows the original characters in the five fonts.
\begin{figure}[t] \begin{center}
    \includegraphics[width=3.3in]{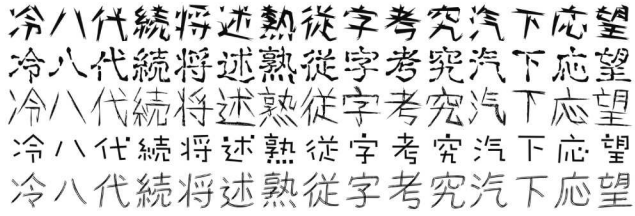}
    \caption{The Ibara, Gcomic, Onryo, Tsunoda, and Zinpen fonts (from top to bottom).} \label{fig:examples}
\end{center} \end{figure}
The characters in each font have distinctive strokes;
for instance, two parallel lines are used in Zinpen.
It is very difficult for existing methods to generate characters in these characteristic fonts.

We created sample images by drawing characters in a 360-pt font size
using the Qt library\footnote{\url{http://www.qt.io/}}.
The size of the images is 500 by 500 pixels.
We implemented the proposed method with C++ and experiments were carried out on a Windows machine with a dual-core CPU.
Fifteen characters were chosen as the samples in each font.
The coefficients were set as follows: $\beta_1=2.0$ and $\beta_2=1.5$.
We generated 2,965 characters, which comprise the set of JIS level-1.

For comparison, we used the existing method \cite{id63} based on patch transformation developed by Saito {\it et al}.
To the best of our knowledge,
\cite{id63} is the only method that is applicable to characters in various fonts.
\cite{id63} divides samples into grid squares and generates characters by deploying the squares.
The frameworks of the proposed and \cite{id63} are similar.
Both exploit samples, extract components of characters, and generate characters.
However, the extracted components are significantly different.
The components in \cite{id63} are small pieces of characters that lack meaning,
whereas the components in the proposed method are complete strokes.

We have employed our method for sample selection, the results of which are summarized in Table \ref{tab:jgg}.
We extracted strokes of the validation characters from the 1,006 characters from kyoiku-kanji
that are for elementary school students, as determined by the Japanese Ministry of Education.
The initial candidates are randomly generated.
We fixed the number of elements in a candidate to 15 characters.
In each iteration of the algorithm, 20 candidates survive as good candidates and 150 new candidates are created.
The maximum number of iterations is set to 1,000.
We varied $\alpha$ and carried out sample selection.
The minimum $f_{\text{\it selection}}$ and $f_r$ are listed in Table \ref{tab:jgg}.
The obtained samples are complex characters at high values of $\alpha$ and simpler at low $\alpha$ values.
We use samples at $\alpha = 0.6$ as the sample characters in the following experiments.
\begin{table}[t] \begin{center}
    \caption{Sample selection results.} \label{tab:jgg}
    \begin{tabular}{|c|c|c|c|} \hline
      $\alpha$ &$f_{\text{\it selection}}$ &$f_r$ &samples \\ \hline
      0.8 &0.017 &0.673 &\begin{minipage}{50mm} \centering
        \includegraphics[width=45mm]{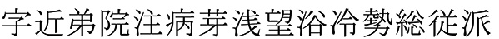} \end{minipage} \\ \hline
      0.6 &0.027 &0.447 &\begin{minipage}{50mm} \centering
        \includegraphics[width=45mm]{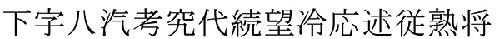} \end{minipage} \\ \hline
      0.4 &0.030 &0.387 &\begin{minipage}{50mm} \centering
        \includegraphics[width=45mm]{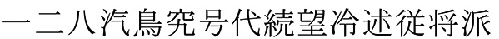} \end{minipage} \\ \hline
    \end{tabular}
\end{center} \end{table}

\subsection{Character generation results}
\begin{figure}[t] \begin{center}
    \begin{tabular}{c}
      \includegraphics[width=3.3in]{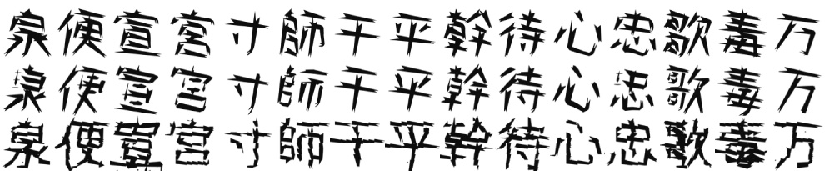}\\ (a) \\
      \includegraphics[width=3.3in]{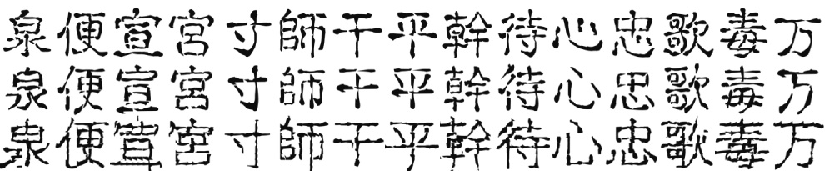}\\ (b) \\
      \includegraphics[width=3.3in]{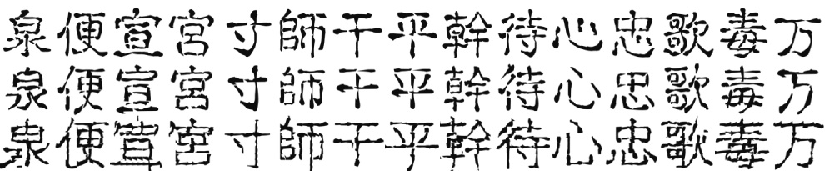}\\ (c) \\
      \includegraphics[width=3.3in]{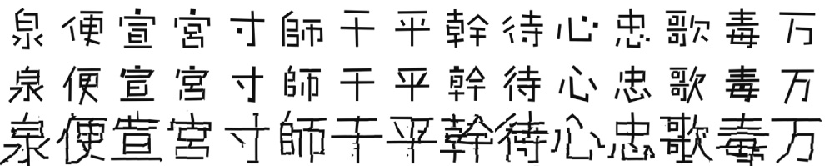}\\ (d) \\
      \includegraphics[width=3.3in]{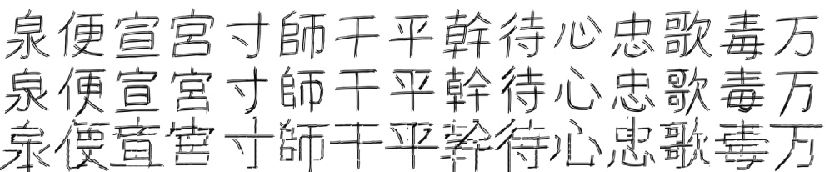}\\ (e)
    \end{tabular}
    \caption{
      The characters generated in five fonts:
      (a) Ibara, (b) Gcomic, (c) Onryo, (d) Tsunoda, and (e) Zinpen.
      From top to bottom in each subfigure, results are shown for the original characters,
      the proposed method, and the method of Saito {\em et al.} \cite{id63}.
    }\label{fig:generated}
\end{center} \end{figure}

We generated characters in the five fonts using the proposed and existing methods,
which are shown in Fig. \ref{fig:generated}.
We generated 2,965 characters in each font, though only 75 characters are shown due to space limitations.
Almost all characters generated by the proposed method appear clean and have good readability.
Moreover, the font characteristics, such as the slant of the characters in Ibara, are successfully reconstructed.
It is not easy to distinguish the original characters from those generated by the proposed method at a glance.
\textcolor{black}{
  In addition, we describe the complexity of the strokes of the generated characters:
  1 (min), 29 (max), 12.5 (avg), and 4.5 (std).
}

\subsection{Similarity evaluation of the images}
We evaluated the generated characters as images.
We compared the generated characters with the original images with the Chamfer distance\cite{borgefors1988hierarchical}.
Let $d_{\text{\it cham}}(A,B)$ be the Chamfer distance from image $A$ to image $B$.
The distance is defined as $d_{\text{\it cham}}(A,B) = \sum_{p\in \hat{A}} \min_{q \in \hat{B}} |p - q|$,
where $\hat{A}$ and $\hat{B}$ are the sets of edge points of $A$ and $B$, respectively.
We obtained $\hat{A}$ and $\hat{B}$ by applying a Canny edge detector \cite{canny}.
In general, $d_{\text{\it cham}}(A,B)$ is not equal to $d_{\text{\it cham}}(B,A)$;
therefore, the symmetric formulation is often used
\begin{equation}
  d_{\text{\it cham}}'(A,B) = d_{\text{\it cham}}(A,B) + d_{\text{\it cham}}(B,A).
\end{equation}
We employed $d_{\text{\it cham}}'$ in this study.

A subset of the generated characters is used for this evaluation.
Specifically, we used the 1,006 characters mentioned above.
We resized the images of the original and generated characters to 100 by 100 pixels
and calculated the Chamfer distance between the generated characters and the originals.
Table \ref{tab:chamfer} summarizes the average Chamfer distances.
\begin{table}[t] \centering
  \caption{Average Chamfer distance} \label{tab:chamfer}
  \begin{tabular}{|c|c|c|c|c|c|} \hline
    &Ibara & Gcomic & Onryo  & Tsunoda & Zinpen \\ \hline
    Saito \cite{id63} & 4.7 & 4.2 & 4.6 & 6.8 & 3.8 \\ \hline
    Proposed & 4.1 & 3.7 & 4.2 & 4.1 & 3.2 \\ \hline
  \end{tabular}
\end{table}
The proposed method is closer for all five fonts than the existing method.
In particular, the proposed method is greatly superior to the existing method for the Tsunoda font.
Tsunoda has a distinctive skeleton, as can be seen in the spaces between strokes.
Since the proposed method successfully modified the skeleton for Tsunoda,
the generated characters are close to the originals.
These results numerically demonstrate the effectiveness of the proposed method.

\subsection{Evaluation using character recognition}
We evaluated the generated characters by using them as training data for a character recognition system.
The test data are character images of kyoiku-kanji in each font.
Therefore, the number of classes is 1,006.
A simple recognition method with Chamfer distance is used in this study.
We calculated the Chamfer distance between the training and test data
and classified the test data as the nearest character.
Test data are created as images, including one JPEG-compressed character.
We created test data by drawing characters using a 60-pt font size on 100 by 100 pixel images;
the background is white, and the foreground is black.
Table \ref{tab:recog} summarizes the results.
\begin{table}[t] \centering
  \caption{Recognition results (\%)} \label{tab:recog}
  \begin{tabular}{|c|c|c|c|c|c|} \hline
    & Ibara & Gcomic & Onryo & Tsunoda & Zinpen \\ \hline 
    Saito \cite{id63} & 29.0 & 61.5 & 56.5 & 7.4 & 68.8 \\ \hline
    Proposed & 48.3 & 77.4 & 67.7 & 28.6 & 77.4 \\ \hline
  \end{tabular}
\end{table}
The proposed method achieved higher performance than the existing method for all fonts.

\textcolor{black}{
  \subsection{Subjective evaluation}
}
We present the results of two subjective evaluations based on the similarity of each font to the original characters
and the appearance of the generated characters.
Both subjective evaluations were carried out by 14 participants.

The first subjective evaluation is of the similarity between the generated and original characters.
At the beginning of the first subjective evaluation, we showed the original characters to the participants.
Then, 10 idioms consisting of four characters were displayed.
The idioms were made from original characters, characters generated by the proposed method,
or characters generated by the existing method.
We use the mean opinion score (MOS) over the results.
The participants assigned scores ranging from 1 (bad) to 5 (excellent) to the idioms
according to their impression of their similarity.
Table \ref{tab:sub1} summarizes the results.
The proposed method receives higher MOSs than the existing method.
Moreover, the MOSs of the proposed method are relatively close to the originals.
\begin{table}[t] \begin{center}
    \caption{Results of a subjective evaluation of font similarity to original characters} \label{tab:sub1}
    \begin{tabular}{|c|c|c|c|c|c|} \hline
      & Ibara   & Gcomic & Onryo & Tsunoda & Zinpen \\ \hline 
      Originals & 4.5 & 4.8  & 4.5 & 4.9 & 4.9   \\ \hline
      Saito \cite{id63}  & 1.1 & 1.3 & 1.5 & 1.1 & 1.3 \\ \hline
      Proposed  & 4.3 & 4.4 & 4.6 & 4.3 & 4.6  \\ \hline
    \end{tabular}
\end{center} \end{table}

\begin{figure}[t] \begin{center}
    \begin{tabular}{c}
      \includegraphics[width=3.3in]{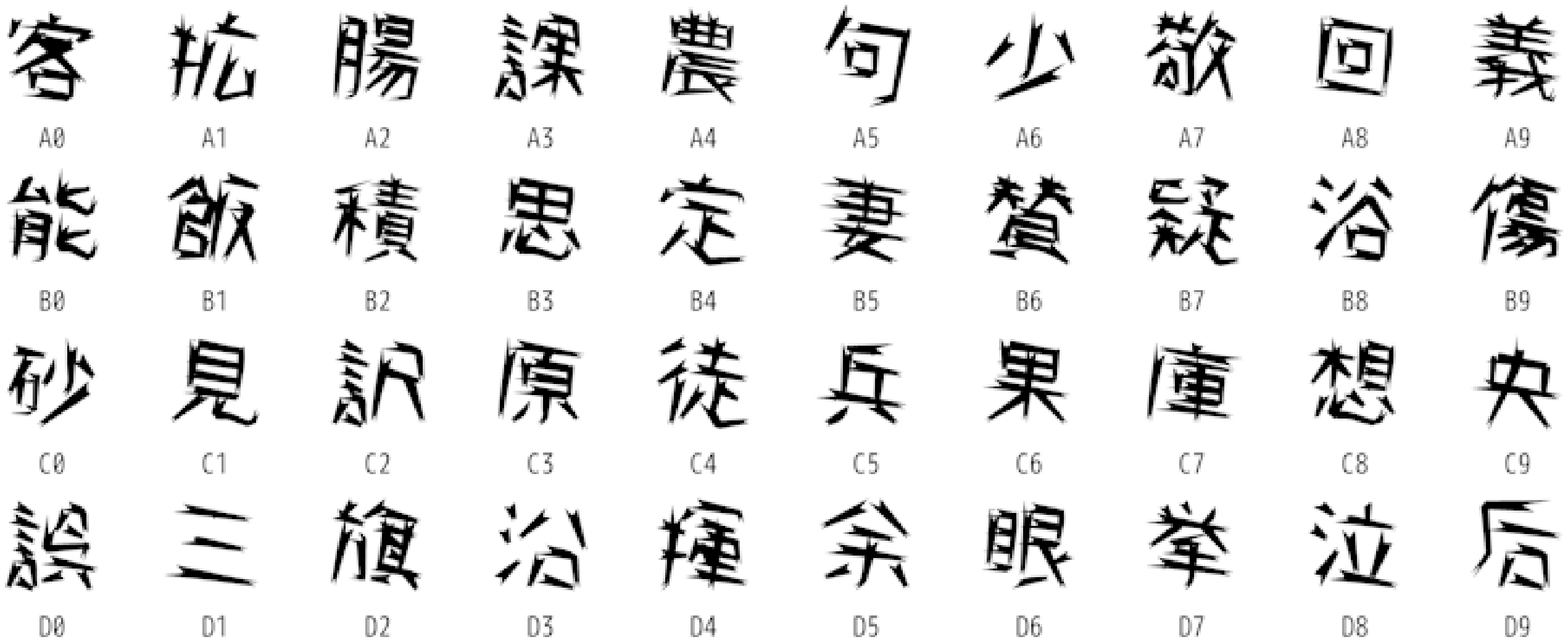}\\ (a) \\
      \includegraphics[width=3.3in]{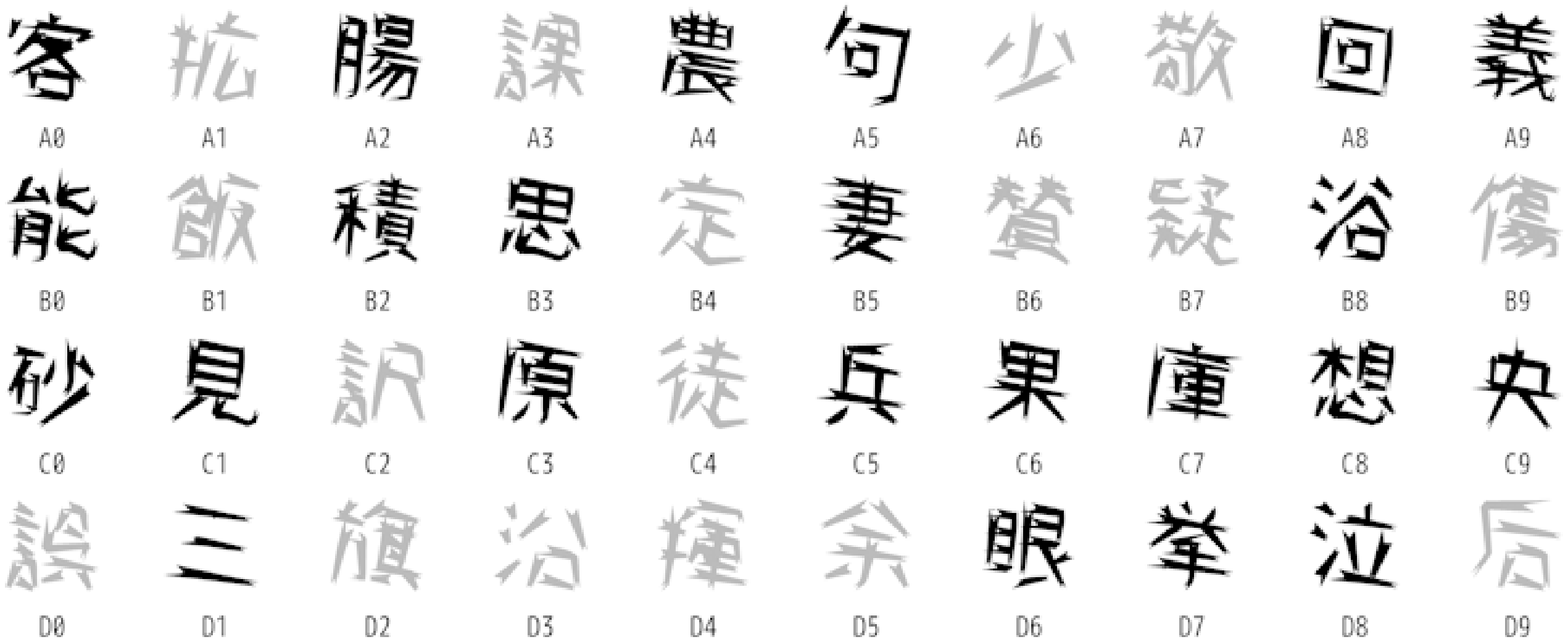}\\ (b)
    \end{tabular}
    \caption{
      Examples of test data and answers.
      (a) Test data shown to the participants.
      (b) Answers to the test data. Generated characters are gray.
    } \label{fig:sub2}
\end{center} \end{figure}

The second subjective evaluation is of appearance.
We asked the participants to select characters that may have been generated by computers.
The test data consisted of 150 characters from each font,
i.e., 50 characters each from the original characters, those generated by the existing method,
and those generated by the proposed method.
Therefore, the test data consist of a total of 750 characters.
Fig. \ref{fig:sub2} illustrates the test data shown to the participants.
We counted the number of participants who believed the characters to have been generated artificially.
Then, we calculated the averages and normalized the numbers.
Table \ref{tab:sub2} summarizes the results.
A character's appearance is natural if the value is low and unnatural if it is high.
{\it Natural} means that the characters are likely handmade, and {\it unnatural} means artificial.
Most of characters generated by the existing method are classified as generated characters
since the results are in the range of 0.77--0.99.
The appearance of the characters generated by the method in \cite{id63} is far from handmade.
In contrast, the results for the proposed method are much lower than those for \cite{id63}.
In particular, the characters generated by the proposed method for Ibara and Onryo are close to the original characters.
According to the results,
it is difficult to distinguish the original characters and those generated by the proposed method, even by a human.
Therefore, the proposed method successfully generated characters with a good appearance.
\begin{table}[t] \begin{center}
    \caption{Results of the subjective evaluation of appearance}  \label{tab:sub2}
    \begin{tabular}{|c|c|c|c|c|c|} \hline
      & Ibara   & Gcomic & Onryo & Tsunoda & Zinpen \\ \hline 
      Originals & 0.06 & 0.02  & 0.04 & 0.01 & 0.01   \\ \hline
      Saito \cite{id63}  & 0.96 & 0.9 & 0.86 & 0.98 & 0.94 \\ \hline
      Proposed  & 0.11 & 0.09 & 0.07 & 0.13 & 0.08  \\ \hline
    \end{tabular}
\end{center} \end{table}

\subsection{Varied font generation}
\begin{figure*}[t] \begin{center}
    \begin{tabular}{cc}
      \includegraphics[width=3.3in]{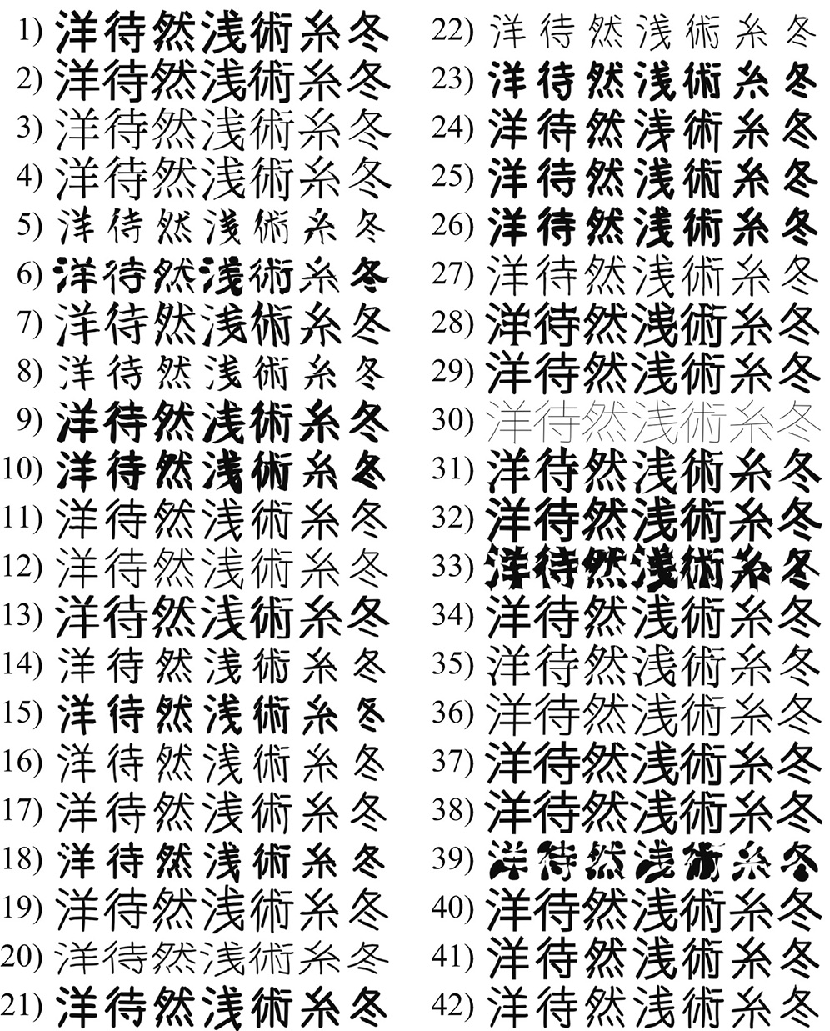}
      &\includegraphics[width=3.3in]{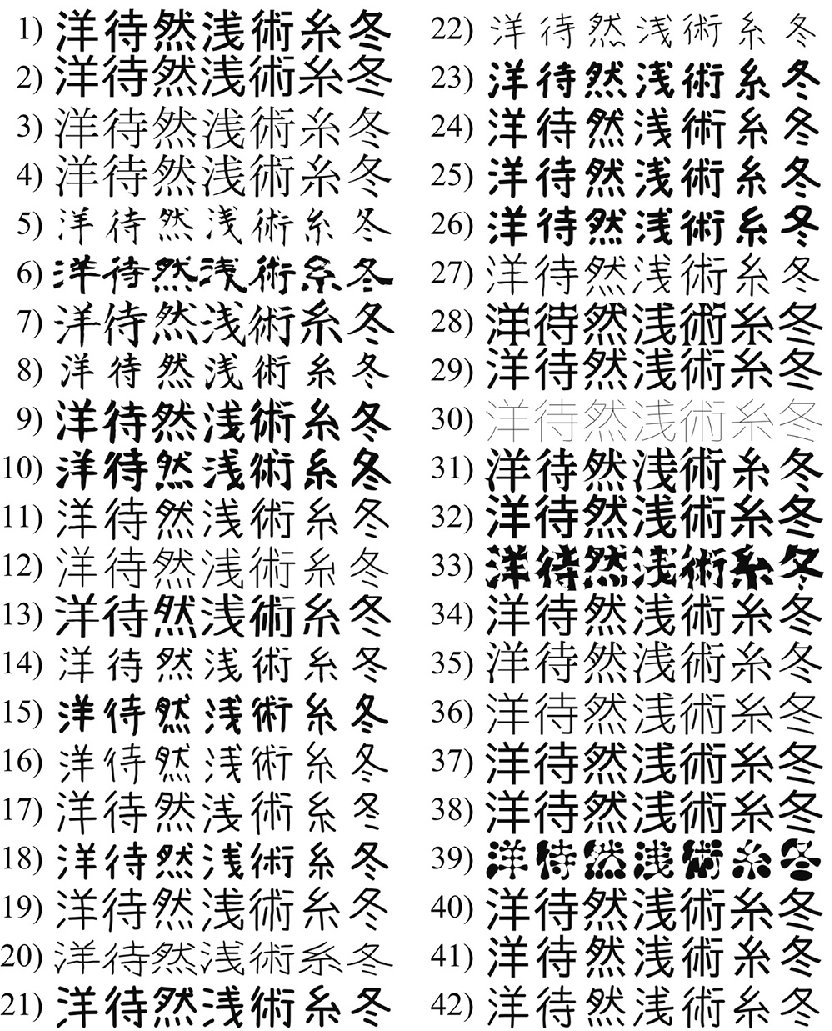} \\
      (a) Proposed &(b) Original
    \end{tabular}    
    \caption{
      Generation results of the proposed method and the original characters in various fonts.
      Each row shows one font.
    } \label{fig:res_various}
\end{center} \end{figure*}

In order to demonstrate the font generation capability of the proposed method,
we performed a generation experiment with 42 fonts:
four standard fonts used in Windows (Gothic, Meiryo, Mincho, and YuMincho),
six calligraphy handwriting fonts \ref{font:Ao}) -- \ref{font:riitf}),
17 pen handwriting fonts \ref{font:azukifont}) -- \ref{font:uzurafont}),
and  15 artificial fonts \ref{font:AMEMUCHIGOTHIC}) -- \ref{font:yutapon-coding}).
The fonts are varied, and include Mincho, Gothic, clerical, antique, personal handwriting,
professional and handwriting styles.
It is worth noting that the number of fonts used in most existing methods is small;
in particular, we used a significantly large number of handwritten fonts, 23.
In this experiment, we generated 2,965 characters.
We illustrate examples of the results in Fig. \ref{fig:res_various}.
The results are promising.
The proposed method generated clean characters and the style of each font is reproduced.

\section{Conclusion} \label{sec:conclusion}
This paper focused on the problem of generating characters for a typographic font
by exploiting a subset of a font and a skeleton dataset.
The proposed method extracts character strokes and constructs characters
by selecting and deploying the strokes to the skeleton.
The proposed method successfully extracts the natural strokes from sample character images.
It is worthwhile to emphasize the importance of stroke extraction from the small number of character images.
The proposed method only requires a font subset---as small as 15 samples---which is a feasible number of samples to collect.
With such a small subset, the proposed method can generate thousands of characters.
Sample collection is further eased because this method can extract characters from image formats.

An experimental evaluation was conducted with five characteristic fonts that are difficult to generate with existing methods.
We evaluated the generated characters by objective and subjective evaluations;
all results indicated that the characters generated by our proposed method have a comparable appearance,
usefulness, and readability to the original characters.
Furthermore, we carried out the experiments with subsets of 42 fonts
to demonstrate the generative capability of the proposed method.
In our future work, we will attempt to automatically adjust skeleton samples and conduct experiments
with a larger number and greater variation of fonts.

\appendix[List of fonts] 
\begin{enumerate}
\item MS Gothic \label{font:MSGothic}
\item MS Meiryo \label{font:MSMeiryo}
\item MS Mincho \label{font:Mincho}
\item MS YuMincho \label{font:YuMincho}
\item \href{http://opentype.jp/aoyagikouzanfontt.htm}{Aoyagi kouzan} \label{font:Ao} 
\item \href{http://opentype.jp/aoyagireisho.htm}{Aoyagi reisho} \label{font:Aorei} 
\item \href{http://www.ac-font.com/jp/detail\_jb\_007.php}{Eishi kaisho} \label{font:Eishi}
\item \href{http://deepblue.opal.ne.jp/faraway/font.html}{aiharahude} \label{font:aiharahude}
\item \href{http://sapphirecrown.xxxxxxxx.jp/}{jetblack} \label{font:jetblack}
\item \href{http://aoirii.babyblue.jp/font/riitf/index.html}{riitf} \label{font:riitf}
\item \href{http://azukifont.com/}{azukifont} \label{font:azukifont}
\item \href{http://welina.holy.jp/font/tegaki/chif/}{chigfont} \label{font:chigfont}
\item \href{http://marusexijaxs.web.fc2.com/tegakifont.html}{gyate} \label{font:gyate-luminescence-tte}
\item \href{http://huwahuwa.ff-design.net/archives/35}{hosofuwafont} \label{font:hosofuwafont}
\item \href{http://ameblo.jp/fakeholic/entry-10972813360.html}{KajudenB} \label{font:KajudenFont-Full-Bold}
\item \href{http://ameblo.jp/fakeholic/entry-10972813360.html}{KajudenR} \label{font:KajudenFont-Full-Regular}
\item \href{http://www.ez0.net/distribution/font/kiloji/}{kiloji} \label{font:kiloji}
\item \href{http://font.spicy-sweet.com/}{KTEGAKI} \label{font:KTEGAKI}
\item \href{http://www.masuseki.com/index.php?u=my\_works/121003\_mitimasu.htm}{mitimasu} \label{font:mitimasu}
\item \href{http://www.geocities.jp/s318shunkasyuto/}{seifuu} \label{font:seifuu}
\item \href{https://osdn.jp/projects/setofont/releases/}{setofont} \label{font:sjis-sp-setofont}
\item \href{http://sana.s12.xrea.com/2\_sanafon.html}{SNsanafon} \label{font:SNsanafon}
\item \href{http://slimedaisuki.blog9.fc2.com/blog-entry-2475.html}{TAKUMISFONT-B} \label{font:TAKUMISFONT-B}
\item \href{http://azukifont.com/}{uzurafont} \label{font:uzurafont}
\item \href{http://www8.plala.or.jp/p\_dolce/font2.html}{apjapanesefonth} \label{font:apjapanesefonth}
\item \href{http://www.kfstudio.net/himaji/}{KFhimaji} \label{font:KFhimaji}
\item \href{http://sapphirecrown.xxxxxxxx.jp/}{ruriiro} \label{font:ruriiro}
\item \href{http://slimedaisuki.blog9.fc2.com/blog-entry-2755.html}{AMEMUCHIGOTHIC} \label{font:AMEMUCHIGOTHIC}
\item \href{http://font.gloomy.jp/honoka-antique-maru-dl.html}{antique} \label{font:antique-maru}
\item \href{http://font.websozai.jp/}{chogokubosogothic} \label{font:chogokubosogothic}
\item \href{https://code.google.com/archive/p/dejima-fonts/}{dejima-mincho} \label{font:dejima-mincho}
\item \href{http://okoneya.jp/font/genei-antique.html}{GenEiAntique} \label{font:GenEiAntique}
\item \href{http://www.ankokukoubou.com/font/hakidame.htm}{hakidame} \label{font:hakidame}
\item \href{https://osdn.jp/projects/ipafonts/}{IPAexg} \label{font:IPAexg}
\item \href{https://osdn.jp/projects/ipafonts/}{IPAexm} \label{font:IPAexm}
\item \href{http://font.cutegirl.jp/jk-font-light.html}{jk-go-l} \label{font:jk-go-l}
\item \href{https://kazesawa.github.io/}{Kazesawa} \label{font:Kazesawa-Regular}
\item \href{http://mix-mplus-ipa.osdn.jp/migu/}{migu-1c} \label{font:migu-1c}
\item \href{http://www2s.biglobe.ne.jp/\textasciitilde fub/font/mofuji.html}{mofuji} \label{font:mofuji}
\item \href{http://itouhiro.hatenablog.com/entry/20140917/font}{Nasu} \label{font:NasuFont}
\item \href{www.flopdesign.com/freefont/smartfont.html}{Smart} \label{font:SmartFontUI}
\item \href{http://net2.system.to/pc/font.html}{yutapon-coding} \label{font:yutapon-coding}
\item \href{http://www2s.biglobe.ne.jp/\textasciitilde fub/font/ibaraji.html}{Ibara} \label{font:ibara}
\item \href{http://material.animehack.jp/font\_gcomickoin.html}{Gcomic} \label{font:gcomic}
\item \href{http://www.ankokukoubou.com/font/onryou.htm}{Onryo} \label{font:onryo}
\item \href{http://a17-s.net/srh/m/f/h694.html}{Tsunoda} \label{font:srtsunoda}
\item \href{http://zinsta.jp/font/freefont.html}{Zinpen} \label{font:zinpen}
\end{enumerate}

\bibliographystyle{IEEEtran}
\bibliography{ref}










\end{document}